\documentclass[times,twocolumn,final]{elsarticle}

\usepackage{medima}
\usepackage{framed,multirow}
\usepackage{soul}
\usepackage[normalem]{ulem}
\useunder{\uline}{\ul}{} 
\usepackage{amssymb} 
\usepackage{latexsym}
\usepackage[switch]{lineno}
\usepackage{pifont}
\usepackage{url}
\usepackage[table,xcdraw]{xcolor}
\usepackage{hyperref}

\definecolor{newcolor}{rgb}{.8,.349,.1}

\journal{Medical Image Analysis} 

\begin{document}
\begin{titlepage}
    \centering
    \vspace*{\fill}
    \includegraphics[width=0.9\textwidth]{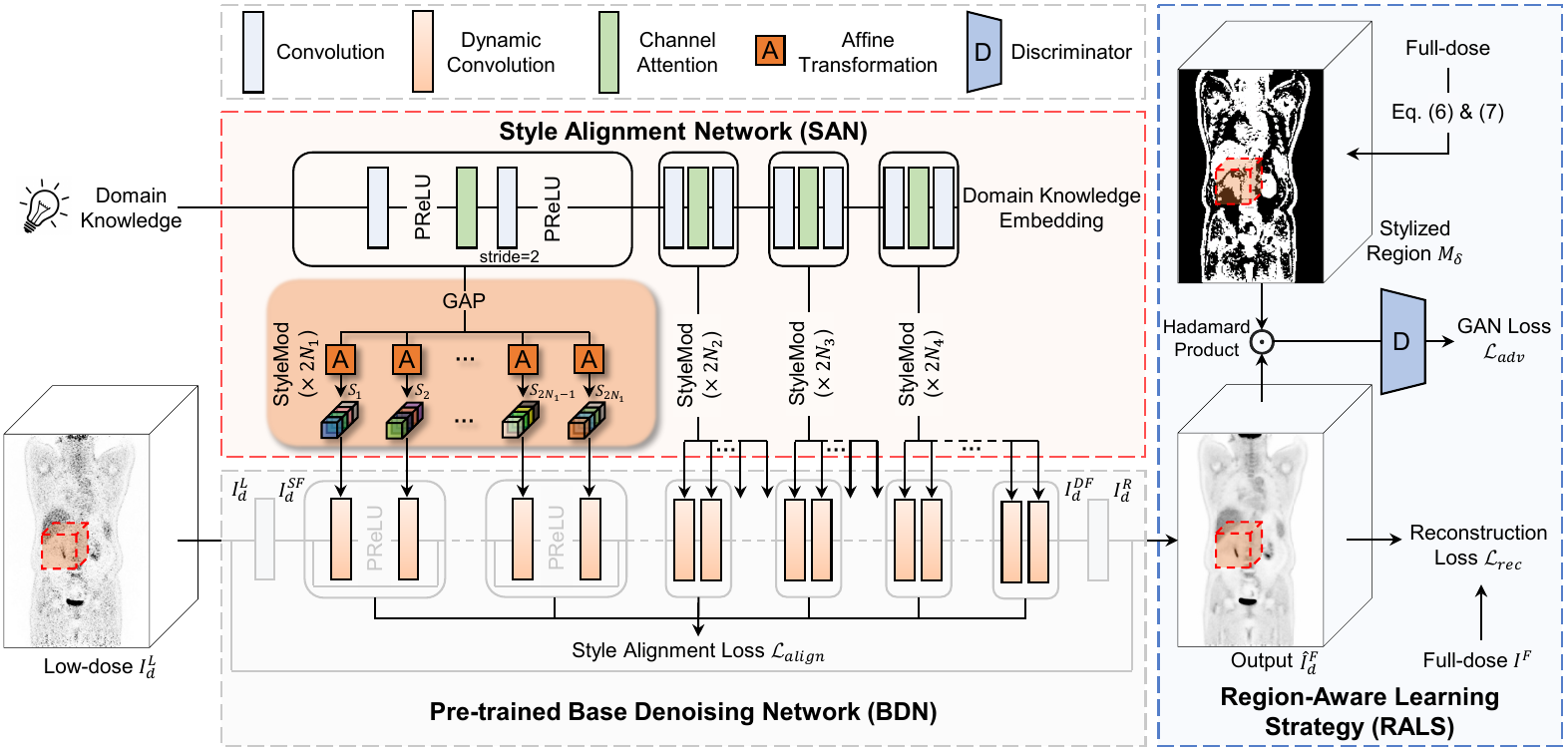}
    \vspace*{\fill}
\end{titlepage}

\verso{Zhiwen Yang \textit{et~al.}}

\begin{frontmatter}

\title{UniPET: a universal network for high-quality PET image denoising across varied dose reduction factors}%



\author[1]{Zhiwen \snm{Yang}}

\author[1]{Yang \snm{Zhou}}

\author[1]{Haowei \snm{Chen}}


\author[2]{Hui \snm{Zhang}}

\author[3]{Dan \snm{Zhao}}

\author[4]{Bingzheng \snm{Wei}}


\author[1]{Yan \snm{Xu} \corref{cor1}}
\ead{xuyan04@gmail.com}
\cortext[cor1]{Corresponding author}

\address[1]{School of Biological Science and Medical Engineering, State Key Laboratory of Software Development Environment, Key Laboratory of Biomechanics, Mechanobiology of Ministry of Education and Beijing Advanced Innovation Centre for Biomedical Engineering, Beihang University, Beijing 100191, China}

\address[2]{Department of Biomedical Engineering, Tsinghua University, Beijing 100084, China} 
\address[3]{Department of Gynecology Oncology, National Cancer Center/National Clinical Research Center for Cancer/Cancer Hospital, Chinese Academy of Medical Sciences and Peking Union Medical College, Beijing 100021, China} 
\address[4]{Xiaomi Corporation, Beijing 100085, China}

\received{1 May 2013}
\finalform{10 May 2013}
\accepted{13 May 2013}
\availableonline{15 May 2013}
\communicated{S. Sarkar}

\begin{abstract}
Most existing deep learning-based PET image denoising methods assume a fixed and known dose reduction factor (DRF) for low-dose PET images. However, these methods encounter significant performance degradation when the DRF varies beyond the assumed one in practical applications. To address the challenge posed by varied DRFs, several preliminary studies focus on the task of universal PET image denoising, aiming to train a universal model over low-dose data across DRFs. Nonetheless, these vanilla universal models often struggle with misaligned styles present in different DRF data, leading to the \textit{style elimination issue} with a significant over-smoothing effect. To deal with this issue, we innovatively introduce domain generalization to PET image denoising and propose a universal PET image denoising network (UniPET) to achieve high-quality PET image denoising across diverse DRFs. UniPET comprises two primary innovations: a style alignment network (SAN) and a region-aware learning strategy (RALS). Specifically, SAN utilizes style alignment techniques derived from domain generalization to align and recover styles across different DRFs, ensuring the model's generalizability across various DRFs while effectively preserving styles. Furthermore, to enhance style recovery, RALS distinguishes between flat and stylized regions, exclusively conducting adversarial learning on the latter, thereby more effectively guiding the model's focus towards learning stylized regions. It is demonstrated that our proposed UniPET can adaptively recover different DRF styles and achieve high-quality PET image denoising across DRFs. Comprehensive experiments show that UniPET exhibits comparable performance to individual DRF-specific models at specific DRFs and realizes state-of-the-art performance in universal PET image denoising quantitatively, perceptually, and clinically. Code is available at \href{https://github.com/Yaziwel/UniPET}{https://github.com/Yaziwel/UniPET}.
\end{abstract}


\end{frontmatter}


\section{Introduction}
\label{sec_intro}
Positron emission tomography (PET) image denoising is a clinically important task focused on recovering high-quality PET images from low-dose counterparts. This process addresses key challenges in nuclear medicine, where the need to reduce radiation exposure to patients through low-dose imaging protocols (\textit{e.g.,} reduced tracer dose or shorter scanning duration) while maintaining diagnostic quality. Due to the limited photon counts received during acquisition, low-dose PET images often suffer from severe noise and reduced signal-to-noise ratio, which compromises their quantitative and diagnostic accuracy. To tackle this issue, recent deep learning-based methods have achieved remarkable success in recovering high-quality PET images from low-dose images \citep{xiang_autocontext_2017,xu_200x_2017,chan_DCNN_2018,wang_3DcGAN_2018,wang_autocontext_GAN_2019,schaefferkoetter_Unet_2020,zhou_cyclegan_2020,Chaudhari_external_evaluation_2021,luo_transformer_GAN_2021,zhou_SGSGAN_2022,zeng_3d_cvtgan_2022,luo_frequency_mia_2022,yang_drmc_2023,fu_AIGAN_2023,jang_spach_transformer_2023,jiang_ran_mia_2023}.

\begin{figure}[t]
\centering
\includegraphics[width=0.45\textwidth]{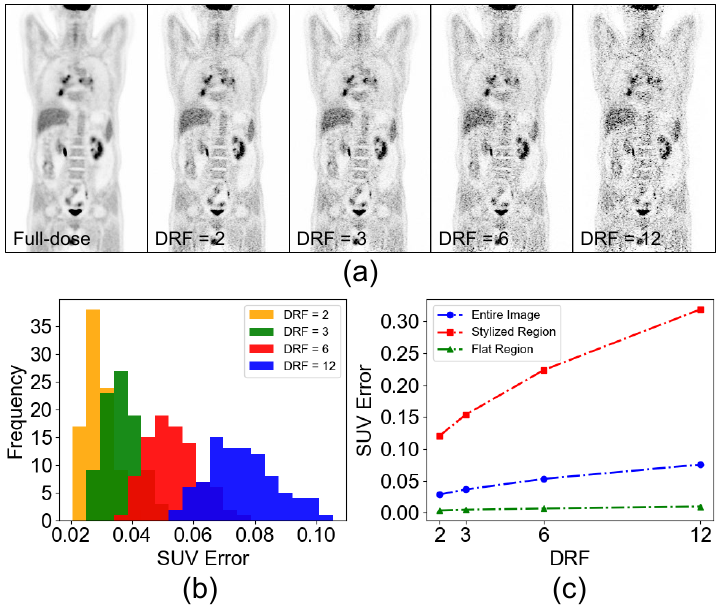}
\caption{Visualization and error analysis of low-dose PET images with varying DRFs. (a) PET images at different DRFs. (b) Histogram of the SUV error distribution for PET images at varying DRFs. (c) SUV error across different regions of the PET images at varying DRFs, including the entire image, stylized regions, and flat regions. The SUV errors in (b) and (c) are measured using the mean absolute error (MAE) on the training set of the UPID-Base dataset.} 
\label{fig_intro_drf_shift}
\end{figure} 

Despite their notable success, most existing deep learning-based PET image denoising methods are customized for low-dose data with specific dose reduction factors (DRFs) and exhibit limited generalizability in practical applications \citep{xiang_autocontext_2017,xu_200x_2017,wang_3DcGAN_2018,wang_autocontext_GAN_2019,zhou_cyclegan_2020,Chaudhari_external_evaluation_2021,luo_transformer_GAN_2021,zhou_SGSGAN_2022,zeng_3d_cvtgan_2022,luo_frequency_mia_2022,fu_AIGAN_2023,jang_spach_transformer_2023,jiang_ran_mia_2023}. This raises two primary concerns. (1) DRF-specific models encounter significant performance degradation when exposed to diverse DRFs in practical applications. These models typically assume that the input DRF is fixed and known, which contrasts with real-world scenarios where DRFs often fluctuate due to variations of the patients' body mass index (BMI) and the associated tracer injection protocols \citep{sanchez2014drf_bmi}, the timing of post-tracer injection acquisition \citep{Xie24_dose_aware_diffusion_jnm}, and the total scanning duration \citep{Xie24_dose_aware_diffusion_jnm}. However, as shown in \autoref{fig_intro_drf_shift} (a), Low-dose PET images at different DRFs exhibit distinct visual characteristics. While they stem from the same underlying clean image (\textit{i.e.,} the full-dose image), each DRF possesses a unique style, with its textures and details degraded by a specific level of noise. Moreover, \autoref{fig_intro_drf_shift} (b) reveals that different DRFs show varying error distributions, indicating that they belong to different data distributions and can be considered distinct domains. Consequently, a DRF-specific model specialized for data from a certain DRF struggles to handle the complex domain shift resulting from DRF variations. (2) Training separate models for each specific DRF from scratch is also impractical due to the considerable demands for computational resources and the inefficiencies in maintenance. Therefore, there is a pressing need to develop a universal model with strong generalizability that can accommodate different DRF inputs.

To simultaneously handle various DRF inputs with a single model, several preliminary studies have undertaken the task of training a universal model over multiple distinct DRF datasets \citep{chan_DCNN_2018,schaefferkoetter_Unet_2020}. By parameter sharing across different DRFs, the universal model can leverage shared knowledge across domains, resulting in consistently robust performance across diverse DRF datasets. 

Nevertheless, when compared to DRF-specific models tailored for each DRF, the vanilla universal model with unified modeling would suffer from the \textit{style elimination issue} — the model tends to eliminate diverse DRF-specific styles (textures and details) in favor of a more generalized output, resulting in over-smoothing effect in the synthesized images \citep{schaefferkoetter_Unet_2020}.  This issue arises when the universal model encounters diverse DRF data with misaligned styles due to varying levels of noise degradation. The universal model,  in an attempt to learn a common representation for all different DRF data, tends to prioritize the most prevalent features while inadvertently eliminating DRF-specific styles. In the context of PET images, styles refer to the textures and details associated with small lesions and subtle spatial patterns. The elimination of these styles significantly impacts lesion detectability and tracer uptake accuracy. The \textit{style elimination issue} highlights the incapacity of a vanilla universal model to faithfully restore the intricate details in PET images of different DRFs. Thus, addressing misaligned styles and mitigating the \textit{style elimination issue} becomes imperative for enhancing the generalizability of vanilla universal models to various DRFs.


Domain generalization (DG) \citep{pan_DG_remove_2018,jin_DG_remove_2020,zhou_DG_remove_2020,wang_DG_align_2021,jeon_DG_remove_2021,ding_DG_remove_2022,Peng_DG_remove_2022,park_DG_remove_2023} provides a potential solution for tackling the \textit{style elimination issue} through style alignment. Considering the fact that styles vary across domains, DG techniques often engage in the delicate art of either removing or aligning styles to obtain domain-invariant features. In the context of PET image denoising, styles serving as the embodiment of textures and details hold paramount importance. The method of removing styles may exacerbate the \textit{style elimination issue} while aligning styles to preserve them can effectively alleviate this problem. In light of this, we embrace the approach of style alignment and propose a novel DG technique to improve the model's generalizability across DRFs.

\begin{figure}[t]
\centering
\includegraphics[width=0.45\textwidth]{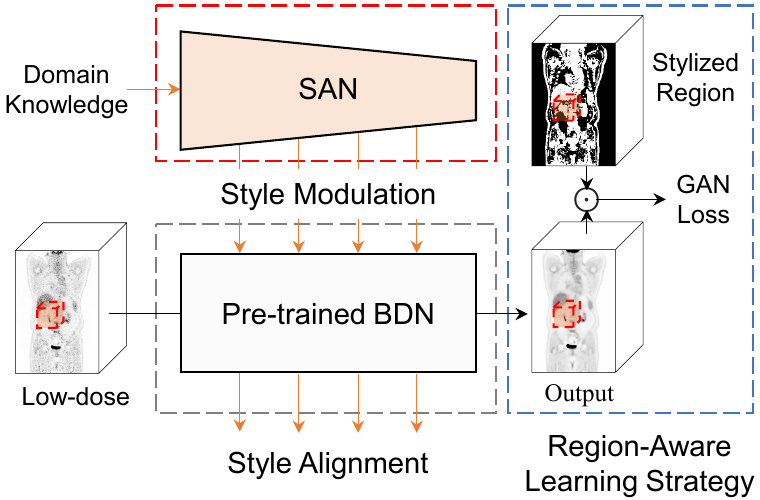}
\caption{A brief overview of UniPET for universal PET image denoising. UniPET mainly consists of a pre-trained base denoising network (BDN), a style alignment network (SAN), and a region-aware learning strategy (RALS).} 
\label{fig_briefview}
\end{figure} 

In this paper, we address \textit{style elimination issue} of vanilla universal models and aim to develop a universal model with strong generalizability to facilitate high-quality PET image denoising across varied DRFs. To accomplish this, we innovatively present a universal PET image denoising network (UniPET, see \autoref{fig_briefview}) which comprises three key components: (1) Base denoising network (BDN): Initially pre-trained on a multi-DRF dataset, BDN aims at producing a preliminary prediction result. Like vanilla universal models, it can yield satisfactory approximation but struggles to recover fine details in PET images with varied DRFs. (2) Style alignment network (SAN): SAN, serving as a DG technique, is introduced to mitigate the \textit{style elimination issue} and refine the style recovery of the BDN. SAN takes domain knowledge as input and dynamically aligns styles of different DRF features. By harmonizing diverse DRF features into a common space, SAN enables generalization across distinct DRFs while preserving style characteristics, resulting in the generation of realistic textures for specific DRFs. (3) Region-aware learning strategy (RALS): RALS is proposed to emphasize the network’s focus on style recovery. This approach is motivated by the observation in \autoref{fig_intro_drf_shift}(c) that the stylized region exhibits the majority of errors in low-dose images and is highly sensitive to DRF variations, while the flat region shows only minor errors and minimal variation. With RALS, our model treats stylized regions and flat regions differently, conducting GAN training \citep{goodfellow_GAN_2014,gulrajani_wgan_gp_2017} exclusively on the former. This separation encourages the model to prioritize style recovery while preventing overfitting on the easily recoverable flat regions. Notably, our proposed UniPET can adopt networks of various architectures as its BDN, indicating its versatility and flexibility. Comprehensive experiments demonstrate that our proposed UniPET achieves comparable performance to individual DRF-specific models at specific DRFs and achieves state-of-the-art performance in universal PET image denoising quantitatively, perceptually, and clinically.

Our contribution can be summarized as follows: 
\begin{itemize}
\item  We present a universal PET image denoising network (UniPET) to achieve high-quality PET image denoising across diverse DRFs. UniPET exhibits strong generalizability in dealing with DRF variations present in practical applications.

\item  We innovatively harness the concept of domain generalization (DG) and propose a style alignment network (SAN) to adaptively recover styles and align different DRF features into a common feature space. SAN ensures both domain generalization and exceptional performance in specific DRFs.

\item  We employ a region-aware learning strategy (RASL), prioritizing style recovery while preventing overfitting on flat regions. This strategy effectively restores intricate details in PET images.
\end{itemize}

\section{Related works}
\subsection{ Universal PET image denoising} 
Positron emission tomography (PET) is a widely used functional imaging technique for disease diagnosis and intervention. It enables the visualization and quantification of \textit{in vivo} metabolic processes by detecting photon emissions from injected radioactive tracers \citep{phelps2000PET_metabolism}. Tissues with higher metabolic rates accumulate more tracer, resulting in stronger photon emissions detected by the PET scanner. However, due to the inherent radiation exposure risk from these tracers, low-dose imaging protocols (such as reducing tracer dosage or scanning time) are highly desirable. Unfortunately, low-dose PET imaging results in limited photon counts during acquisition, leading to increased image noise and a lower signal-to-noise ratio. This compromises both diagnostic quality and quantitative accuracy of the PET images. To address this, PET image denoising is introduced to recover full-dose PET images from low-dose PET images after dose reduction. 

Recently, deep learning-based methods have become the main workhorse to tackle PET image denoising \citep{xiang_autocontext_2017,xu_200x_2017,chan_DCNN_2018,wang_3DcGAN_2018,wang_autocontext_GAN_2019,schaefferkoetter_Unet_2020,zhou_cyclegan_2020,Chaudhari_external_evaluation_2021,luo_transformer_GAN_2021,zhou_SGSGAN_2022,zeng_3d_cvtgan_2022,luo_frequency_mia_2022,yang_drmc_2023,fu_AIGAN_2023,jang_spach_transformer_2023,jiang_ran_mia_2023}. These approaches focus on two key aspects: network designs and loss functions. Network designs often leverage convolutional neural networks (CNNs) \citep{xiang_autocontext_2017,xu_200x_2017,chan_DCNN_2018,wang_3DcGAN_2018,wang_autocontext_GAN_2019,schaefferkoetter_Unet_2020,zhou_cyclegan_2020,Chaudhari_external_evaluation_2021,zhou_SGSGAN_2022,luo_frequency_mia_2022,fu_AIGAN_2023,jiang_ran_mia_2023}, Transformers \citep{luo_transformer_GAN_2021,zeng_3d_cvtgan_2022,yang_drmc_2023,jang_spach_transformer_2023}, or their hybrid variants to ensure optimal mapping. In terms of loss functions, beyond standard $l_1$ and $l_2$ losses for image reconstruction, studies also explore generative adversarial network (GAN) loss \citep{wang_3DcGAN_2018,wang_autocontext_GAN_2019,zhou_cyclegan_2020,zhou_SGSGAN_2022,zeng_3d_cvtgan_2022,luo_frequency_mia_2022,fu_AIGAN_2023}, task-oriented loss \citep{zhou_SGSGAN_2022}, and spectrum constraint \citep{luo_frequency_mia_2022} to refine network optimization and improve denoising quality. Nonetheless, many deep learning methods are tailored to low-dose PET images with specific dose reduction factors (DRFs), like DRF=4 signifying a reduction to 25$\%$ of the original dose. Consequently, these DRF-specific methods will suffer a performance drop in practical applications where the DRF varies beyond the assumed one.  

To enhance model performance across various DRFs, several preliminary studies have pursued universal PET image denoising by training a single model over diverse DRF data \citep{chan_DCNN_2018,schaefferkoetter_Unet_2020}. \citet{chan_DCNN_2018} introduces a deep convolutional neural network (DCNN) with residual learning to adaptively estimate the noise residuals of different DRF data. This approach demonstrates more consistent performance in the liver region across various DRF data compared to DRF-specific models. \citet{schaefferkoetter_Unet_2020} conduct a clinical assessment on a Unet trained over PET images in a wide range of DRFs. Results indicate that the Unet can yield better denoising performance than Guassian smoothing for all DRF data. Nonetheless, the synthesized results demonstrate some degree of blurring and over-smoothing. These preliminary studies solely rely on the network's mapping capabilities to handle low-dose data from different DRFs. However, they do not address the style misalignment among different DRF data, where the texture and detail vary with DRF due to differing levels of noise caused by reduced photon counts \citep{sanaei2021_detail_vary_with_drf,vogel2023_texture_vary_with_drf}. A unified processing without adaptively tackling the misaligned styles will result in the emergence of the \textit{style elimination issue} within the universal model. In this paper, we tackle the universal PET image denoising problem from a domain generalization perspective and propose two modules, SAN and RALS, to alleviate the \textit{style elimination issue}.

\subsection{Domain generalization} 
The goal of domain generalization (DG) is to learn domain-invariant features that can be transferable across domains \citep{pan_DG_remove_2018,jin_DG_remove_2020,zhou_DG_remove_2020,wang_DG_align_2021,jeon_DG_remove_2021,ding_DG_remove_2022,Peng_DG_remove_2022,park_DG_remove_2023}. In recent DG studies, image features are typically disentangled into domain-invariant content and domain-specific styles. The domain-invariant content typically refers to the semantic information of the image, while domain-specific styles refer to the textures and details present in the image. To reduce the discrepancy between features of different domains, DG studies often involve removing or aligning domain-specific styles to acquire domain-invariant features. In tasks such as segmentation \citep{Peng_DG_remove_2022} and classification \citep{pan_DG_remove_2018,jin_DG_remove_2020,zhou_DG_remove_2020,jeon_DG_remove_2021,ding_DG_remove_2022}, where semantic content plays a vital role and style information is less critical, DG approaches tend to eliminate style information while preserving domain-invariant semantic content. However, in tasks like image restoration \citep{wang_DG_align_2021}, where style information significantly influences outcomes, DG studies often try to align various domain styles so that different domain features are projected into a common space. In PET image denoising, style that represents textures and details is essential for accurate results. Texture reflects the spatial arrangement and heterogeneity of uptake values within a region. Different organs and tissues show characteristic spatial heterogeneity driven by biology: the liver typically has a smooth, relatively uniform SUV distribution, whereas the brain displays structured heterogeneity that follows anatomy (higher uptake in gray matter than in white matter). Textures are clinically important because they have been shown to correlate with tumor grade and treatment response \citep{pineiro2021PET_texture}. Detail refers to the high-frequency components of the image, highlighting small-scale structural variations and edge sharpness that are crucial for detecting small lesions. In short, texture is about the pattern and variation of SUV values in a region (spatial heterogeneity), while detail in PET is about how finely you can see and resolve structures (spatial resolution, sharpness). Both texture and detail (\textit{i.e.,} style) severely influence the overall diagnostic quality. Therefore, we adopt the style alignment approach to facilitate the model’s generalizability across data with different DRFs while preserving styles.

\subsection{Style modulation} 
Style modulation emerges from the style transfer literature \citep{huang_mod_in_2017} and typically employs extra auxiliary information as guidance to manipulate styles of network features \citep{huang_mod_in_2017,chen_mod_BN_2018,karras_stylegan_2019,karras_stylegan2_2020}. The style information in CNNs is encapsulated within the mean and standard deviation of extracted features \citep{huang_mod_in_2017}. Therefore, style modulation generally estimates denormalization factors (e.g., bias and scaling) based on auxiliary information and applies them to normalized features to manipulate their styles (i.e., mean and standard deviation). As a result, these denormalized features acquire new attributes from the auxiliary information. Typical style modulation techniques are based on normalization layers like instance normalization (IN) \citep{huang_mod_in_2017,karras_stylegan_2019}, batch normalization (BN) \citep{chen_mod_BN_2018}, and weight normalization (WN) \citep{karras_stylegan2_2020}. \citet{huang_mod_in_2017} utilizes IN-based modulation to render one image using style information from another image. \citet{karras_stylegan_2019} achieve precise control over styles in image generation tasks with IN-based style modulation techniques. In their subsequent research \citep{karras_stylegan2_2020}, they replace IN with WN, resulting in finer style modulation effects and the generation of images with more realistic textures. Motivated by the promising performance of modulation mechanisms in image generation, we employ a WN-based style modulation mechanism to align and recover styles in various DRF PET images. 


\begin{figure*}[t]
\centering
\includegraphics[width=0.9\textwidth]{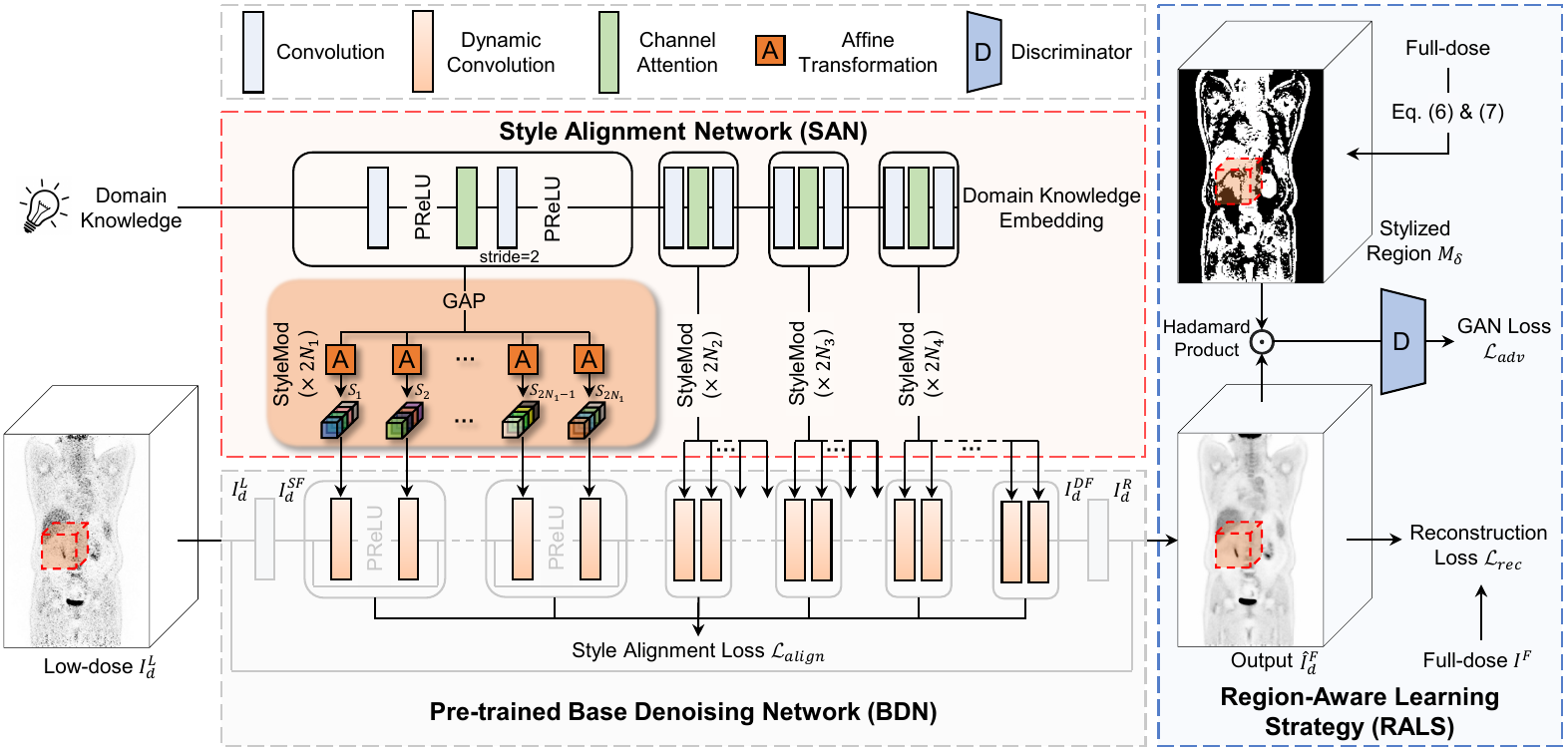}
\caption{The framework of UniPET for universal PET image denoising. UniPET comprises a pre-trained base denoising network (BDN), a style alignment network (SAN), and a region-aware learning strategy (RALS). BDN is proposed for coarse-grained denoising. It adopts a residual learning structure with a convolution layer for shallow feature $I_d^{SF}$ extraction, a sequence of $N=\sum_{i=1}^{4}N_{i}$ residual blocks for deep feature $I_d^{DF}$ extraction, and a convolution layer for image residual $I_d^{R}$ estimation. SAN focuses on fine-grained style alignment and recovery across varied DRFs. SAN consists of 4 encoding blocks to extract 4 distinct hierarchical embeddings, which are then used to recover styles of different hierarchical features (4 groups, each with $N_i$ residual blocks) within BDN through style modulation. RASL is introduced to direct the model's focus towards learning stylized regions. For the sake of simplicity, the illustration of the PReLU activation \citep{he2015PReLU} in the subsequent blocks of BDN and SAN has been omitted.
} 
\label{fig_overview}
\end{figure*} 

\subsection{Region-aware learning} 
Region-aware learning is an efficient strategy that directs the network to focus on regions of interest (ROIs) \citep{qin2019difficulty,xie_frequency_2021,zhou_SGSGAN_2022,jiang_ran_mia_2023}
. Typically, an image is divided into ROIs and others. After region division, ROIs are assigned more computational resources for more complex processing. \citet{qin2019difficulty} introduces a difficulty-aware network that categorizes images into regions based on their difficulty and performs complex operations specifically on high-difficulty regions. \citet{xie_frequency_2021} propose a frequency-aware network that decomposes each image into high-frequency, mid-frequency, and low-frequency regions according to its coefficients in the discrete cosine transform (DCT) domain. These regions are dynamically allocated computation resources based on their frequency. \citet{zhou_SGSGAN_2022} directly employ the segmentation of ROIs as guidance to preserve these ROIs with the help of downstream segmentation tasks. These studies successfully prioritize the preservation of ROIs by assigning more computational resources. In this paper, we emphasize the recovery of styles by introducing a novel region-aware learning strategy that exclusively applies GAN training to stylized regions.

\section{Methodology} 
In the universal PET image denoising task, our objective is to train a single universal model capable of restoring low-dose PET images across different DRFs. \autoref{fig_overview} presents a detailed illustration of our proposed UniPET, consisting of three key components: (1) a pre-trained base denoising network (BDN) producing estimation results from different DRF inputs; (2) a style alignment network (SAN) dedicated to aligning distinct DRF styles and refining the style recovery process of BDN; (3) a region-aware learning strategy (RALS) designed to enhance BDN's focus on stylized regions. Specifically, BDN takes a low-dose PET image as input. Initially, the preliminary estimations from BDN are suboptimal and lack style information. SAN and RALS are subsequently introduced to fine-tune the style recovery of BDN, aiming to restore textures and details, including small lesions and subtle spatial patterns. SAN uses domain knowledge to guide BDN's style recovery, adaptively aligning different domain styles into a common feature space, thereby enhancing BDN's generalizability and preserving style information. RALS treats flat and stylized regions differently, exclusively conducting GAN training on stylized regions, allowing BDN to concentrate more on style recovery. The details of our proposed UniPET are elaborated in the following.

\subsection{Base denoising network} 
The base denoising network (BDN) acts as the backbone for universal PET image denoising. As is shown in \autoref{fig_overview}, BDN employs a residual learning structure to estimate the residuals of the input. Given a low-dose PET image $I^{L}_{d}$ of a specific DRF $d$, where $d \in \left[d_{min}, d_{max}\right]$, BDN first applies a $K\times K\times K$ convolution layer to obtain shallow feature $I^{SF}_{d}$. These shallow features undergo refinement and deep feature extraction through $N$ residual blocks, resulting in the deep feature $I^{DF}_{d}$. Eventually, another $K\times K\times K$ convolution layer generates the estimated residual feature $I^{R}_{d}$. The synthesized image is then produced by element-wise sum $\hat{I}_{d}^{F}=I^{L}_{d} + I^{R}_{d}$. It's important to note that BDN is pre-trained on a multi-DRF dataset, enabling it with coarse-grained denoising capabilities.

\subsection{Style alignment network} 
Although BDN possesses basic denoising abilities, it encounters the \textit{style elimination issue} stemming from its inability to handle misaligned styles across DRFs. To overcome this challenge, we leverage the style alignment technique in domain generalization and introduce a style alignment network (SAN). SAN is designed to dynamically align diverse domain styles and unify distinct domain features into a common space. This approach ensures the model's generalizability while preserving the fidelity of styles. SAN primarily operates through three key steps: (1) domain knowledge embedding, which encodes domain information into embeddings, offering valuable guidance for style alignment; (2) style modulation, utilizing the encoded domain knowledge embedding as guidance to align processed features' styles in BDN through modulation operations; (3) style alignment loss, serving as the objective function for style alignment. We will delve into a comprehensive explanation of these three steps in the subsequent paragraphs.

\textbf{(1) Domain knowledge embedding.} 
The role of domain knowledge is to guide the model in perceiving domain-related information and adaptively processing inputs based on the specific input domain. Domain knowledge should inform domain variations and integrate effectively into the model. Typical domain knowledge encompasses the input image itself, as well as other domain-informative features extracted from it. Specifically, the input image partially reflects the characteristics of the input domain and varies with domain changes. But it also contains a significant amount of domain-invariant information, which makes it less domain-informative. To derive more effective domain knowledge, handcrafted features can be extracted from the input image to filter out irrelevant or less domain-informative information. For example, the high-frequency component (HFC) \citep{liu_promt_evp_2023} of the input image removes low-frequency information—typically less sensitive to domain variations—and preserves domain-specific high-frequency details. This makes the HFC a potentially more effective representation of domain knowledge. Nevertheless, both the input image and handcrafted features still lack the adaptability needed to integrate into the model for effective guidance. In this study, we demonstrate that shallow features $I_d^{SF}$ represent a simple yet effective form of domain knowledge. On the one hand, shallow features are adaptively derived from the input image through a simple learnable convolution, capturing low-level patterns such as textures and details \citep{zeiler2014cnnvisualizing}. These low-level patterns have been reported to be highly sensitive to domain variations and are commonly used to model domain discrepancies \citep{li2020domain_low_level, cardace2022shallow_feature}. On the other hand, shallow features serve as the initial stage of deep feature extraction, aligning closely with subsequent layers, which facilitates the provision of feature-level guidance for the model.

After acquiring the domain knowledge, we encode it into different hierarchical embeddings. We deploy a straightforward convolutional neural network with $T$ blocks to encode and interpret the domain-specific information. Within each block, the convolution layer uses a kernel size of $K$, while the final convolution layer employs a strided convolution for downsampling. Furthermore, to enhance domain sensitivity, channel attention layers are incorporated to capture inter-channel dependencies. As a result, the domain knowledge undergoes encoding at four distinct hierarchical levels, yielding embeddings with unique hierarchies. These embeddings play a crucial guiding role in the subsequent style recovery process for different hierarchical features within BDN.

\textbf{(2) Style modulation.} With the extracted domain knowledge embeddings, our objective is to harmonize various DRF styles within BDN into a common space. A critical challenge lies in establishing interactions between these embeddings and the styles in need of manipulation within BDN. In recent advancements, StyleGAN \citep{karras_stylegan_2019,karras_stylegan2_2020} has seen remarkable success in image generation, precisely controlling style and detail generation through style modulation. Drawing inspiration from StyleGAN, we introduce style modulation to manipulate and interact with styles within BDN. Specifically, we first derive a style code $S \in \mathbb{R}^{C}$ (C denotes the channel number) from the domain knowledge embedding $E$ using the equation:
    \begin{equation}
    \centering
    S=\textbf{A}(\textbf{GAP}(E)),
    \end{equation}
where $\textbf{GAP}(\cdot)$ signifies the global average pooling layer, and $\textbf{A}(\cdot)$ represents the corresponding affine transformation layer, involving multiplication by a weight matrix and addition of a bias. Subsequently, the style code is integrated into the respective convolution layer within BDN for style modulation. This integration turns the standard convolution layer into a dynamic convolution through the following equation:
\begin{equation}
W^{\prime}=\frac{W * S}{\sqrt{\sum_{i, j}(W * S)^2+\epsilon}},
\end{equation}
where $W$ and $W^{\prime}$ represent the weights in the standard convolution and after style modulation, respectively. $i$ and $j$ enumerate the input features and spatial footprint of the convolution, respectively. In contrast to the standard convolution, where the parameter and the processed feature remain independent, dynamic convolution exhibits the capability to dynamically interact with features and modulate feature styles.

To achieve more precise style modulation, we divide the $N$ residual blocks in BDN into $T$ groups sequentially, where $N=\sum_{i=1}^{T}N_{i}$, with each group comprising $N_i$ residual blocks, respectively. Finally, we utilize the $T$ different hierarchical domain knowledge embeddings generated by SAN to individually modulate the styles of these $T$ groups of residual blocks. This methodology enables SAN to perform more accurate and flexible style modulation on different hierarchical features within BDN.

\textbf{(3) Style alignment loss.} The objective of style modulation is to mitigate the style discrepancy and align different DRF styles. As outlined in paper \citep{huang_mod_in_2017}, in CNNs, the style information is encapsulated in the channel-wise mean and standard deviation of each feature. Consequently, the style of the output feature $I_{d}^{B_iF}$ at a specific DRF $d$ of the $i$-th residual block $B_i$ within BDN can be expressed through the concatenation of the channel-wise mean ${\mu}_d^{B_i} \in \mathbb{R}^{C}$ and standard deviation ${\sigma}_d^{B_i}  \in \mathbb{R}^{C}$ statistics, which is denoted as $\Phi_d^{B_i} \in \mathbb{R}^{2C}$:
\begin{equation}
\Phi_d^{B_i}=\left[\mu_d^{B_i}, \sigma_d^{B_i}\right].
\end{equation} 

The collective style at DRF $d$ is then represented as the concatenation of all styles $\Phi_d^{B_i}$ derived from the output features across $N$ residual blocks, which is denoted as $\Phi_{d} \in \mathbb{R}^{2 \cdot N \cdot C}$:
\begin{equation}
\Phi_d=\left[\Phi_d^{B_1}, \Phi_d^{B_2} \ldots, \Phi_d^{B_N}\right].
\end{equation} 

The discrepancy in style between different DRF data can be formulated as $\left\|\Phi_{d_j}-\Phi_{d_k}\right\|_1$, where $d_j, d_k \in [d_{min},d_{max}]$. However, directly minimizing this discrepancy involves the cumbersome process of loading identical patient data across various DRFs, leading to time-consuming procedures and uncertain convergence in the final style space. To address this, we turn to use full-dose data at DRF $d=1$ to generate a reference full-dose style and then minimize the discrepancy between the low-dose style $\Phi_{d}$ and the reference full-dose style $\Phi_{1}$. The generation of the full-dose style is reasonable as reconstructing full-dose data at DRF $d=1$ is also one of our goals to recover different DRF data. Consequently, we introduce the following style alignment loss to diminish the style discrepancy:
\begin{equation}
\mathcal{L}_{align} = \left\|\Phi_{d}-\Phi_{1}\right\|_1.
\end{equation} 

With $\mathcal{L}_{align}$, the style modulation dynamically adjusts various DRF styles to converge towards the full-dose style, thereby aligning these styles within a common space. This approach ensures effective style recovery and bolsters the model's ability to generalize across different DRFs.

\subsection{Region-aware learning strategy} 
It's important to note that both the denoising process in BDN and the style modulation process in SAN are learned over the entire image. However, the majority of PET images consist of simple and flat regions, with only a small portion containing rich style information that presents challenges in denoising. Such data imbalance hampers the network's ability to effectively learn style information. While some studies highlight the efficacy of GAN-based \citep{wang_3DcGAN_2018,wang_autocontext_GAN_2019} methods in texture and style recovery, GANs also require a substantial amount of data during training \citep{karras_gan_limited_2020}. When facing such data imbalance, GANs are prone to overfitting in flat regions. To alleviate this data imbalance and enable the network to better capture style information, we propose a region-aware learning strategy (RALS). RALS directs the network's focus toward stylized regions, facilitating improved mitigation of style discrepancies across DRFs and enhancing style recovery. In this section, we will present RALS by elucidating (1) the identification of stylized regions and (2) region-aware adversarial learning.

\textbf{(1) Identification of stylized regions.} The stylized region within the PET image is characterized as the area rich in textures, while the flat region is defined by its uniformity and minimal texture content. In order to distinguish between stylized and flat regions within the image, we first measure the richness of texture information by calculating the variance around each voxel. Specifically, we densely extract image patches for each spatial location $(i,j,k)$ of the full-dose PET image $I^F$ and compute the variance map $V$ using a local window:
\begin{equation}
V(i, j, k)=\textbf{VAR}\left(\left\{I^F(i+l, j+m, k+n)\right\}_{l, m, n \in\{-1,0,1\}}\right),
\end{equation}
where $\textbf{VAR}(\cdot)$ denotes the variance function applied to the window centered at $(i,j,k)$. Then, a binary mask $M_{\delta}$, representing the stylized region, can be derived by applying a threshold $\delta$:
\begin{equation}
M_\delta(i, j, k)=\left\{\begin{array}{l}
1, V(i, j, k) \geq \delta \\
0, V(i, j, k)<\delta.
\end{array}\right.
\end{equation}

\textbf{(2) Region-aware adversarial learning.} After acquiring the stylized region, we incorporate it into the GAN framework, capitalizing on GAN's strong proficiency in style recovery. Specifically, the denoising result $\hat{I}_d^F$ and the corresponding full-dose data $I^F$ are multiplied by the same mask $M_{\delta}$ to acquire their stylized region content $\hat{I}_{d, M_\delta}^F$ and $I_{M_\delta}^F$, respectively:
\begin{equation}
\hat{I}_{d, M_\delta}^F=\hat{I}_d^F \odot M_\delta, \quad I_{M_\delta}^F=I^F \odot M_\delta .
\end{equation} 
Finally, $\hat{I}_{d, M_\delta}^F$ and $I_{M_\delta}^F$  are fed into the discriminator $\textbf{D}$ for adversarial learning. We employ the WGAN-GP \citep{gulrajani_wgan_gp_2017} as our GAN training scheme, utilizing a discriminator $\textbf{D}$ akin to PatchGAN \citep{isola_pix2pix_2017}. The generator $\textbf{G}$ in our UniPET is comprised of SAN and BDN. The overall adversarial loss $\mathcal{L}_{adv}$ is a composition of the generator's loss $\mathcal{L}_{adv}^{G}$ and the discriminator's loss $\mathcal{L}_{adv}^{D}$:
\begin{equation}
\begin{array}{r}
\mathcal{L}_{a d v}^D =\mathrm{\textbf{E}}_{z \sim \mathbb{O}}[\textbf{D}(z)]-\mathrm{\textbf{E}}_{z \sim \mathbb{P}}[\textbf{D}(z)], \\
 +\lambda \mathrm{\textbf{E}}_{z \sim \mathbb{Q}}\left[\left(\left\|\nabla_z \textbf{D}(z)\right\|_2-1\right)^2\right]
\end{array}
\end{equation}

\begin{equation}
\mathcal{L}_{a d v}^G =-\mathrm{\textbf{E}}_{z \sim \mathbb{O}}[\textbf{D}(z)],
\end{equation}
where $\lambda$ denotes the weighting parameter, $\mathbb{O}$ denotes distribution of masked synthesized data $\hat{I}_{d, M_\delta}^F$, $\mathbb{P}$ denotes distribution of masked full-dose data $I_{M_\delta}^F$, $\mathbb{Q}$ denotes sampling distribution on the straight line between $\mathbb{O}$ and $\mathbb{P}$. The second term of $\mathcal{L}^{D}_{adv}$ is to enforce the Lipschitz constraint. The GAN training process adversarially aligns the synthesized distribution $\mathbb{O}$ with the full-dose distribution $\mathbb{P}$.

During this process, we exclusively conduct GAN training within the stylized region, allowing the network to focus more on learning style recovery. It's noteworthy that the selection of the stylized region aligns with the theory of GANs, as we haven't altered the fundamental workings of GANs but rather narrowed down GAN's training data from the entire image to the stylized region.

\begin{table*}[t] 
\caption{Dataset information.}
\resizebox{\textwidth}{!}{
\begin{tabular}{cccccccccccc}
\hline
Type                     & Dataset         & Institution                           & System                 & Tracer       & Dose    & Low-dose DRF          & Spacing (mm$^3$) & Shape            & Training Patient & Validation Patient & Testing Patient \\ \hline
\multirow{3}{*}{Private} & UPID-Base       & Peking Union Medical College Hospital & PolarStar m660         & $^{18}$F-FDG & 369 MBq & 2,3,6,12              & 3.15×3.15×1.87   & 192×192×$slices$ & 90               & 10                 & 15              \\
                         & UPID-OOD-DRF    & Peking Union Medical College Hospital & PolarStar m660         & $^{18}$F-FDG & 359 MBq & 1.5,2.4,4,10          & 3.15×3.15×1.87   & 192×192×$slices$ & —                & —                  & 15              \\
                         & UPID-OOD-Center & Beijing Hospital                      & PolarStar Flight       & $^{18}$F-FDG & 354 MBq & 2,3,4                 & 3.12×3.12×1.75   & 192×192×$slices$ & —                & —                  & 15              \\ \hline
Public                   & Bern            & University Hospital of Bern           & Biograph Vision Quadra & $^{18}$F-FDG & 219 MBq & 2, 4, 10, 20, 50, 100 & 1.65×1.65×1.65   & 440×440×644      & 295              & 32                 & 50              \\ \hline
\end{tabular}}
\label{table_dataset}
\end{table*}

\subsection{Objective function}
The objective function of UniPET mainly consists of the image reconstruction loss $\mathcal{L}_{rec}$, the style alignment loss $\mathcal{L}_{align}$ in SAN, and the adversarial loss $\mathcal{L}_{adv}$ in RALS. We choose $l_1$ distance as $\mathcal{L}_{rec}$ to constrain the basic voxel-wise image content reconstruction:
\begin{equation}
\mathcal{L}_{rec} = \left\|\hat{I}_d^F-I^F\right\|_1.
\end{equation} 

The overall objective function can be formulated as follows:
\begin{equation}
\mathcal{L} = \mathcal{L}_{rec} + \beta\mathcal{L}_{align} + \gamma\mathcal{L}_{adv},
\end{equation} 
where $\beta$ and $\gamma$ are balancing parameters.
\section{Experimental setup} 
\subsection{Dataset} 
To evaluate the effectiveness and generalizability of the proposed UniPET for universal PET image denoising (UPID), we have established four distinct whole-body datasets. These include three private datasets—UPID-Base, UPID-OOD-DRF, and UPID-OOD-Center—and one publicly available dataset, Bern \citep{xue2022berndataset}. The UPID-Base dataset is used for model training, validation, and testing. The UPID-OOD-DRF and UPID-OOD-Center datasets are both utilized for out-of-distribution (OOD) testing of the model trained on UPID-Base. The UPID-OOD-DRF dataset uses low-dose PET images with unknown DRFs, while the UPID-OOD-Center dataset employs low-dose PET images from an unknown center. The publicly available Bern dataset is utilized for model training, validation, and testing on publicly accessible data. A brief overview of these datasets is provided in Table 1, with detailed descriptions of each dataset presented below. 

\textbf{(1) UPID-Base Dataset.} The UPID-Base dataset is a private dataset utilized for model training, validation, and testing. It contains 115 subjects from the Peking Union Medical College Hospital, with each subject administered an average injection dose of 369 MBq of $^{18}$F-FDG tracer. All data are acquired using a PolarStar m660 PET/CT system in list mode. The post-injection tracer uptake time is 60 minutes and the scan duration is 3 minutes. We store the raw list mode data and then generate simulated low-dose list mode data through random list mode decimation, based on predefined DRFs of 2, 3, 6, and 12. Both full-dose PET images (DRF=1) and low-dose PET images (DRF=2, 3, 6, 12) are reconstructed from the corresponding list mode data using the standard OSEM method. The shape of the reconstructed image is 192 $\times$ 192 $\times$ $slices$ with a voxel size of 3.15 $\times$ 3.15 $\times$ 1.87 mm$^3$. The 115 subjects are divided into 90 for training, 10 for validation, and 15 for testing. Among them, there are 45 subjects with cancers (12 lung, 9 breast, 9 liver, 7 colorectal, 5 lymphoma, and 3 pancreatic cancers), which are divided into 33 for training, 2 for validation, and 10 for testing.

\textbf{(2) UPID-OOD-DRF Dataset.} The UPID-OOD-DRF dataset is a private dataset utilized for OOD testing of the model trained on UPID-Base, utilizing low-dose images with unknown DRFs. It contains 15 additional subjects, all collected from the same center as the UPID-Base dataset. The primary differences lie in the DRFs of the low-dose PET images and the average injection dose. The DRFs for the low-dose images are 1.5, 2.4, 4, and 10, differing from those in the UPID-Base dataset to enable model testing with unknown DRFs. The average injection dose of the $^{18}$F-FDG tracer for the 15 subjects is 359 MBq.  All other imaging procedures and image characteristics remain consistent with those of the UPID-Base dataset.

\textbf{(3) UPID-OOD-Center Dataset.} The UPID-OOD-Center dataset is a private dataset utilized for OOD testing of the model trained on UPID-Base, utilizing low-dose images from an unknown center. It contains 15 subjects from the Beijing Hospital, with each subject administered an average injection dose of 354 MBq of $^{18}$F-FDG tracer. All data are acquired using a PolarStar Flight PET/CT system in list mode. The post-injection tracer uptake time is 60 minutes and the scan duration is 2 minutes. We store the raw list mode data and then generate simulated low-dose list mode data through random list mode decimation, based on predefined DRFs of 2, 3, and 4. Both full-dose PET images (DRF=1) and low-dose PET images (DRF=2, 3, 6, 12) are reconstructed from the corresponding list mode data using the standard OSEM method. The shape of the reconstructed image is 192 $\times$ 192 $\times$ $slices$ with a voxel size of 3.12 $\times$ 3.12 $\times$ 1.75 mm$^3$. 

\textbf{(4) Bern Dataset.} The Bern dataset \citep{xue2022berndataset} is a publicly available dataset used for model training, validation, and testing. It includes 377 subjects from the University Hospital of Bern, with each subject administered an average injection dose of 219 MBq of $^{18}$F-FDG tracer. All data are acquired using a Siemens Biograph Vision Quadra PET/CT system. The post-injection tracer uptake time is 60 minutes and the scan duration is 10 minutes. The dataset includes full-dose PET images and corresponding low-dose PET images with DRFs of 2, 4, 10, 20, 50, and 100. Each image has dimensions of 440 $\times$ 440 $\times$ 644 with a voxel size of 1.65 $\times$ 1.65 $\times$ 1.65 mm$^3$. The 377 subjects are divided into 295 for training, 32 for validation, and 50 for testing.

\subsection{Evaluation metrics} 
\textbf{(1) Quantity evaluation.} We select the commonly used peak signal to noise ratio (PSNR) and structural similarity (SSIM) to assess the denoising performance. PSNR evaluates the accuracy of overall intensity recovery, with a higher PSNR indicating better recovery of the overall SUV value in PET images. SSIM measures structural similarity, with a higher SSIM indicating better preservation of image structures and greater consistency.

\textbf{(2) Perception evaluation.} We utilize the learned perceptual image patch similarity (LPIPS) \citep{zhang_lpips_2018} to evaluate image perceptual quality. Unlike PSNR and SSIM, LPIPS is highly sensitive to perceptual variations like textures, structures, and patterns. This characteristic makes LPIPS an effective evaluation metric for assessing style recovery in PET images. LPIPS, on the other hand, evaluates the perceptual quality and the recovery of textural details. A lower LPIPS score signifies superior restoration of image textures, enhancing the clarity of lesions and spatial patterns—a critical factor for improving lesion detectability in diagnostic contexts.

Additionally, following the paper \citep{li_sacnn_2020}, a comparative reader score is conducted using the UPID-Base testing dataset, among which 10 out of 15 patients have lesions. Specifically, four radiologists independently evaluate the perceptual quality of the synthesized images produced by six different methods applied to low-dose images with DRF=12. In total, each radiologist reviews 90 images (15 synthesized images from each of six methods). They provide scores for noise reduction (image noise level), structure preservation (anatomy edges), texture preservation (spatial patterns and small lesions), and overall quality on a five-point scale, where 1 denotes unacceptable and 5 represents excellent performance.

\textbf{(3) Clinical task evaluation.} According to the RELAINCE guidelines \citep{jha2022RELAINCE_guidelines} and its companion paper \citep{bradshaw2022nuclear_depolyment}, clinical task evaluations are essential to verify a method’s clinical utility. To this end, we assess our model’s performance through a series of clinical task evaluations. We measure the standard uptake value (SUV) errors between synthesized and full-dose images in three regions of interest (ROIs): blood pool, liver, and lesion. The lesion is particularly significant as it represents a crucial area for disease diagnosis. The liver and blood pool are large, metabolically active areas in the body, and their SUV values serve as background references for the metabolic activity of other tissues \citep{hofheinz2016bloodpoolliverlesion}. For example, to achieve a more accurate disease assessment, radiologists often compare lesion SUV values with those of the liver or blood pool for normalization according to previous studies \citep{hofheinz2016bloodpoolliverlesion}. Therefore, it is also important to recover SUV values from these two useful reference regions.  We employ a senior radiologist to annotate these three ROIs across the entire UPID-Base dataset. We utilize the mean absolute error (MAE) to quantify the SUV error (\textit{i.e.,} the SUV difference between predicted and full‑dose images) across three ROIs. Lower MAE values correspond to smaller SUV errors within each ROI, indicating better recovery in those regions.

In addition,  we conduct a lesion detection evaluation task by radiologists to assess the clinical utility of our method. Specifically, we first use the lesions annotated by the senior radiologist on full‑dose PET images as the ground truth. Next, three additional radiologists independently review all PET images under evaluation—including low‑dose PET images and the denoised outputs of various methods—and mark all visible lesions. Finally, we evaluate lesion detection performance using the F1‑score, defined as the harmonic mean of precision and recall. A higher F1‑score indicates improved lesion detectability and, consequently, better diagnostic image quality. This evaluation is conducted on the testing set of the UPID-Base dataset, which includes 15 subjects (5 healthy controls and 10 patients), with the 10 patients collectively presenting 36 lesions. Among the 36 lesions, 11 are located in the lymph node, 9 in the lung, 5 in the liver, 3 in the colorectum, 3 in the breast, 3 in the bone, and 2 in the pancreas. 

Finally, we also compare the lesion detectability of different methods through receiver operating characteristic curves (ROC) derived from a patch‐level lesion classification model. The key insight is that if a denoising method yields restored patches of similar quality as full-dose images, it should achieve comparable lesion detectability with the same classifier, and its ROC curve will lie close to that of the full-dose data. To evaluate this, we extract non-overlapping $64 \times 64 \times 64$ patches from the full-dose PET images in the UPID-Base dataset, excluding any patches that contain only background. Since we have lesion annotations by the radiologist and thus each patch can be labeled as "lesion" or "normal". Following the approach of \citep{cui2023tridoformer}, we train a CNN-based classification model to distinguish between normal and lesion patches, achieving an accuracy of 91\%. We then apply this classifier to patches restored by various denoising methods and plot their ROC curves. The closer a method’s ROC curve is to that of the full‐dose data, the better its image quality and lesion detectability. 

\textbf{(4) Statistical significance testing.} According to the paper \citep{christodoulou2025false_claim}, the lack of statistical significance testing can produce false claims of outperformance in many deep-learning studies. Therefore, we perform paired t-tests for each evaluation metric to compare methods. The nominal significance threshold is set to $\alpha=0.05$. To control the group-wise error rate, we apply a Bonferroni correction within each comparison group by dividing $\alpha$ by the number of comparisons in that group. Comparisons with $p$-values below the corrected threshold are regarded as statistically significant.

\subsection{Implementation details of UniPET} 
\label{sec:implement_UniPET}
\textbf{(1) Network architecture.} Without explicit specification, the convolution kernel size in UniPET is set to $K=3$. In SAN, the number of blocks is set to $T=4$, with channel numbers for these 4 blocks configured as [64, 128, 256, 256]. In BDN, each residual block has a channel number of $C=64$ for all residual blocks, and the total number of residual blocks is set to $N=8$. These 8 residual blocks are divided into $T=4$ groups for style modulation, utilizing the $T=4$ different hierarchical domain knowledge embeddings generated by SAN. Each group consists of $N_1=N_2=N_3=N_4=2$ residual blocks.

\textbf{(2) Data preparation.} All PET images are converted from voxel values to standardized uptake values (SUVs), and both model training, testing, and evaluation are performed in the SUV domain. We train our proposed UniPET on the UPID-Base dataset. The training input for UniPET comprises data from DRFs 1, 2, 3, 6, and 12, where the low-dose data from DRFs 2, 3, 6, and 12 are repeated three times in each training epoch. During training, PET data are clipped to the range [0, 20] and normalized to [0, 1] by dividing by 20 following the paper \citep{aksu2025norm_method}. A batch size of four patches is used, with each patch randomly extracted from a distinct patient to ensure patient-level diversity. Extracted patches undergo an automated validation process to exclude non-informative regions: patches containing only air (i.e., devoid of anatomical structures) are discarded, triggering re-sampling from the same patient until a valid patch is obtained.  During the testing phase, the input whole-body PET image is divided into non-overlapping patches for model processing, after which the synthesized patches are simply merged to reconstruct the whole-body image.

\textbf{(3) Training configuration.} We implement our proposed UniPET in Pytorch on a workstation with NVIDIA A100 GPU. We utilize the Adam optimizer with a fixed learning rate of 0.0001. Balancing parameters in loss functions are set as $\beta=0.001$ and $\gamma=0.001$. The batch size is set as 4. The threshold to determine the stylized region is set to $\delta = 0.001$ In the training process,  we initially exclude SAN and deactivate the style modulation by setting $S=[1, 1, ...,1]\in \mathbb{R}^{64}$ and pre-train BDN using $\mathcal{L}_{rec}$ for 200 epochs as no significant improvement afterward. Subsequently, we activate SAN for style modulation and optimize the entire generator $G$, composed of SAN and BDN, using $\mathcal{L}_{rec}$ and $\mathcal{L}_{align}$ for 200 epochs. Finally, incorporating RALS, we alternate training between $G$ and $D$ for 100 epochs. 


For hyperparameter selection, we employ a group sequential optimization strategy \citep{wang2025grouped_sequential_optimization_strategy}, in which hyperparameters are first partitioned into groups by their estimated impact on final performance and then tuned sequentially from highest to lowest impact: (1) \textbf{data‑related} patch size $P$, which determines the amount of usable input information; (2) \textbf{model‑related} residual block number $N$ and channel dimension $C$, which govern model capacity; (3) \textbf{loss-related} balancing factors $\beta$ and $\gamma$ and the stylized region threshold $\delta$, which control the scale of auxiliary objective for fine-detail recovery. Prior work \citep{liang2021swinir,liang2022details} shows that data‐ and model‐related hyperparameters have a large‐scale impact on overall restoration performance, whereas auxiliary losses primarily affect local details. Accordingly, we tune the groups in the order data → model → loss. Specifically, we first define a candidate value set for each hyperparameter through empirical analysis, then initialize them according to the principle of model simplification—hyperparameters influencing model complexity are initialized to minimal values, while those that influence the training loss are initialized to values that disable the corresponding loss terms. The order in which hyperparameters are tuned is as follows: $P \in \{32, 48, 64, 96\}$, $N \in \{4, 8, 12, 16\}$, $C \in \{16, 32, 64, 128\}$, $\beta \in \{0, 0.0001, 0.001, 0.01, 0.1\}$, $\gamma \in \{0, 0.0001, 0.001, 0.01, 0.1\}$, and $\delta \in \{0, 0.0001, 0.001, 0.01, 0.1\}$. For each hyperparameter, we consider both model performance on the validation dataset and model efficiency. This group sequential optimization strategy offers two key advantages: (1) it achieves faster convergence and significantly reduces computational cost compared to conventional grid search, and (2) it establishes a true coarse‑to‑fine pipeline for high-quality PET image denoising—first ensuring robust global performance by tuning data and model hyperparameters, then focusing on tuning auxiliary loss related hyperparameters to refine the recovery of details.

\begin{table*}[!t] 
\caption{Comparisons between universal and DRF-specific models on the UPID-Base dataset. The four DRF-specific models are individually trained on specific DRF data, whereas the universal model is trained on the entire dataset encompassing all DRFs. Note that DRF-specific models utilize the same network architecture and optimization configurations as UniPET. The best results are highlighted in bold, while the second-best results are underlined. * Denotes results that are significantly different from the best results, based on a paired t-test with Bonferroni correction.}
\centering
\resizebox{\textwidth}{!}{
\begin{tabular}{cc|ccccc|ccccc|ccccc}
\hline
\multicolumn{2}{c|}{\multirow{2}{*}{Method}} & \multicolumn{5}{c|}{PSNR↑}                                                         & \multicolumn{5}{c|}{SSIM↑}                                                              & \multicolumn{5}{c}{LPIPS↓}                                                              \\ \cline{3-17} 
\multicolumn{2}{c|}{}                        & DRF=2          & DRF=3          & DRF=6          & DRF=12         & Avg            & DRF=2           & DRF=3           & DRF=6           & DRF=12          & Avg             & DRF=2           & DRF=3           & DRF=6           & DRF=12          & Avg             \\ \hline
\multicolumn{1}{l|}{}             & DRF=2    & 50.54 $^{*}$        & 48.75 $^{*}$          & 45.64 $^{*}$          & 42.19 $^{*}$          & 46.78 $^{*}$          & \textbf{0.983 \textcolor{white}{$^{*}$}} & 0.975 $^{*}$          & 0.953 $^{*}$          & 0.911 $^{*}$          & 0.955 $^{*}$          & \textbf{0.004 \textcolor{white}{$^{*}$}} & {\ul 0.006 $^{*}$}    & 0.016 $^{*}$          & 0.042 $^{*}$          & 0.017 $^{*}$          \\
\multicolumn{1}{l|}{DRF-specific} & DRF=3    & 50.23 $^{*}$          & {\ul 49.62 \textcolor{white}{$^{*}$}}    & 46.41 $^{*}$          & 43.13 $^{*}$          & 47.35 $^{*}$          & 0.981 $^{*}$          & \textbf{0.977 \textcolor{white}{$^{*}$}} & 0.957 $^{*}$          & 0.917 $^{*}$          & 0.958 $^{*}$          & 0.005 $^{*}$          & \textbf{0.004 \textcolor{white}{$^{*}$}} & 0.012 $^{*}$          & 0.033 $^{*}$          & 0.013 $^{*}$          \\
\multicolumn{1}{c|}{Model}        & DRF=6    & 50.01 $^{*}$          & 49.04 $^{*}$          & {\ul 47.28 $^{*}$}          & 44.33 $^{*}$          & 47.67 $^{*}$          & 0.977 $^{*}$          & 0.973 $^{*}$          & {\ul 0.963 \textcolor{white}{$^{*}$}}    & 0.934 $^{*}$          & 0.962 $^{*}$          & 0.011 $^{*}$          & 0.009 $^{*}$          & \textbf{0.007 \textcolor{white}{$^{*}$}} & 0.015 $^{*}$          & {\ul 0.011 $^{*}$}    \\
\multicolumn{1}{l|}{}             & DRF=12   & 48.84 $^{*}$          & 48.39 $^{*}$          & 47.19 $^{*}$          & 45.03 $^{*}$          & 47.36 $^{*}$          & 0.964 $^{*}$          & 0.961 $^{*}$          & 0.956 $^{*}$          & \textbf{0.947 \textcolor{white}{$^{*}$}} & 0.957 $^{*}$          & 0.020 $^{*}$          & 0.017 $^{*}$          & 0.012 $^{*}$          & \textbf{0.011 \textcolor{white}{$^{*}$}} & 0.015 $^{*}$          \\ \hline
\multicolumn{1}{c|}{Universal}    & BDN & {\ul 50.69 $^{*}$}    & 49.27 $^{*}$          & {\ul 47.28 $^{*}$}    & {\ul 45.08 $^{*}$}    & {\ul 48.08 $^{*}$}    & 0.981 $^{*}$          & 0.975 $^{*}$          & 0.961 $^{*}$          & 0.931 $^{*}$          & {\ul 0.962 $^{*}$}    & 0.011 $^{*}$          & 0.012 $^{*}$          & 0.018 $^{*}$          & 0.024 $^{*}$          & 0.016 $^{*}$          \\
\multicolumn{1}{c|}{Model}        & UniPET (ours)     & \textbf{51.54 \textcolor{white}{$^{*}$}} & \textbf{49.66 \textcolor{white}{$^{*}$}} & \textbf{47.55 \textcolor{white}{$^{*}$}} & \textbf{45.45 \textcolor{white}{$^{*}$}} & \textbf{48.55 \textcolor{white}{$^{*}$}} & {\ul 0.982 $^{*}$}    & {\ul 0.975 $^{*}$}    & \textbf{0.963 \textcolor{white}{$^{*}$}} & {\ul 0.947 \textcolor{white}{$^{*}$}}    & \textbf{0.967 \textcolor{white}{$^{*}$}} & {\ul 0.004 \textcolor{white}{$^{*}$}}    & 0.006 $^{*}$          & {\ul 0.008 $^{*}$}    & {\ul 0.011 \textcolor{white}{$^{*}$}}    & \textbf{0.007 \textcolor{white}{$^{*}$}} \\ \hline
\end{tabular}
} 
\label{table_universal_vs_specific}
\end{table*}

\begin{table}[!t] 
\caption{Model specifications of comparison methods.}
\centering
\resizebox{0.5\textwidth}{!}{
\begin{tabular}{c|c|c|c|c}
\hline
Method            & Params    & Type        & Architecture & GAN                   \\ \hline
Unet              & 128.27 M & CNN         & UNet         &  \\
DCNN              & 3.99 M   & CNN         & ResNet       &  \\
mDCSRN            & 0.41 M   & CNN         & DenseNet     &  \\
3D-cGAN           & 53.62 M  & CNN         & UNet         & $\checkmark$ \\
Spach Transformer & 19.22 M  & Transformer & UNet         &  \\
UniPET            & 2.75 M   & CNN         & ResNet       & $\checkmark$ \\ \hline
\end{tabular}
} 
\label{table_model_info}
\end{table}

\subsection{Implementation details of ablation study} 
To investigate the significance of individual components in
UniPET, we conduct ablation studies on the UPID-Base
dataset specifically focusing on the
style alignment network (SAN), the region-aware learning
strategy (RALS), as well as some key hyperparameters. The implementation details are as follows.

\textbf{(1) Effect of SAN and RALS.} We first perform component analysis on SAN and RALS by disabling one or both components. To disable SAN, we abandon style modulation through setting $S=[1, 1, ...,1]\in \mathbb{R}^{64}$ and discard the style alignment loss $\mathcal{L}_{align}$. To disable RALS, we omit the adversarial loss $\mathcal{L}_{adv}$. To further investigate the effectiveness of these two components, we apply SAN and RALS to other CNN architectures, including ResNet \citep{he_resnet_2016}, DenseNet \citep{huang_densenet_2017}, and UNet \citep{ronneberger_unet_2015}. In this process, the model’s blocks are divided into four groups, each receiving one of the four hierarchical domain knowledge embeddings from SAN. The $3 \times 3$ convolutions within the model blocks are replaced with dynamic convolutions that incorporate style modulation. Both the style alignment loss $\mathcal{L}_{align}$ and the adversarial loss $\mathcal{L}_{adv}$ are applied as well. 

\textbf{(2) Ablation study on SAN.} We conduct ablation studies on key components of SAN, including the types of domain knowledge, the embedding method, and the style alignment loss $\mathcal{L}_{align}$. For the ablation study on domain knowledge, we compare three different types of domain knowledge embeddings: the low-dose image $I_d^L$ itself, the handcrafted high-frequency component (HFC) of $I_d^L$, and our automatically extracted shallow feature $I_d^{SF}$. The HFC is extracted from the input image following the methodology described in \citep{liu_promt_evp_2023}. This process involves transforming the input image to the frequency domain, zeroing out the central low-frequency region, and then applying an inverse transform to return to the image domain. As a result, the HFC retains only the high-frequency components, which are more sensitive to domain variations compared to $I_d^L$. For the ablation study on the domain knowledge embedding method, we compare our hierarchical embedding approach with the single embedding method used in StyleGAN \citep{karras_stylegan2_2020}. Specifically, our hierarchical embedding method utilizes multiple domain knowledge embeddings extracted from different encoding blocks of SAN, while the single embedding method uses only the output embedding of the final block in SAN. For the ablation study on the style alignment loss $\mathcal{L}_{align}$, we compare model performance with and without this loss during training.

\textbf{(3) Ablation study on RALS.} The performance of RALS is primarily influenced by the threshold parameter $\delta$, which defines the stylized region for GAN training. We empirically select a set of candidate values for $\delta \in \{0, 0.0001, 0.001, 0.01, 0.1, +\infty\}$. Two special cases are notable: when $\delta = 0$, GAN training is applied to the entire image, and RALS effectively becomes a standard GAN; when $\delta = +\infty$, no region is selected for GAN training, effectively disabling GAN training during the process. Since RALS involves conducting GAN training on selected regions, we also explore how performing GAN training on different regions—such as the entire image, the stylized region defined by $\delta$, no region, or three specific clinical ROIs—affects performance on those clinical ROIs. In the special case of training GANs on clinical ROIs, we fine-tune the final UniPET model by randomly extracting patches containing the ROI (using available annotations for the three clinical ROIs) from patient data and train the model for 100 epochs.

\textbf{(4) Ablation study on hyperparameter.} We explore the impact of several key hyperparameters on the model's performance, ensuring that when conducting experiments, all settings remain identical to the final UniPET configuration, except for the hyperparameter being explored. Specifically, we modify the value of one hyperparameter at a time, while keeping all others fixed. The tuning sets for each hyperparameter are the same as those used in hyperparameter selection, as outlined in \autoref{sec:implement_UniPET}: patch size $P \in \{32, 48, 64, 96\}$, number of residual blocks $N \in \{4, 8, 12, 16\}$, channel dimension $C \in \{16, 32, 64, 128\}$, loss balancing factor $\beta \in \{0, 0.0001, 0.001, 0.01, 0.1\}$, loss balancing factor $\gamma \in \{0, 0.0001, 0.001, 0.01, 0.1\}$, and the threshold $\delta$ determining the stylized region, where $\delta \in \{0, 0.0001, 0.001, 0.01, 0.1\}$. 

\begin{figure}[t]
\centering
\includegraphics[width=0.5\textwidth]{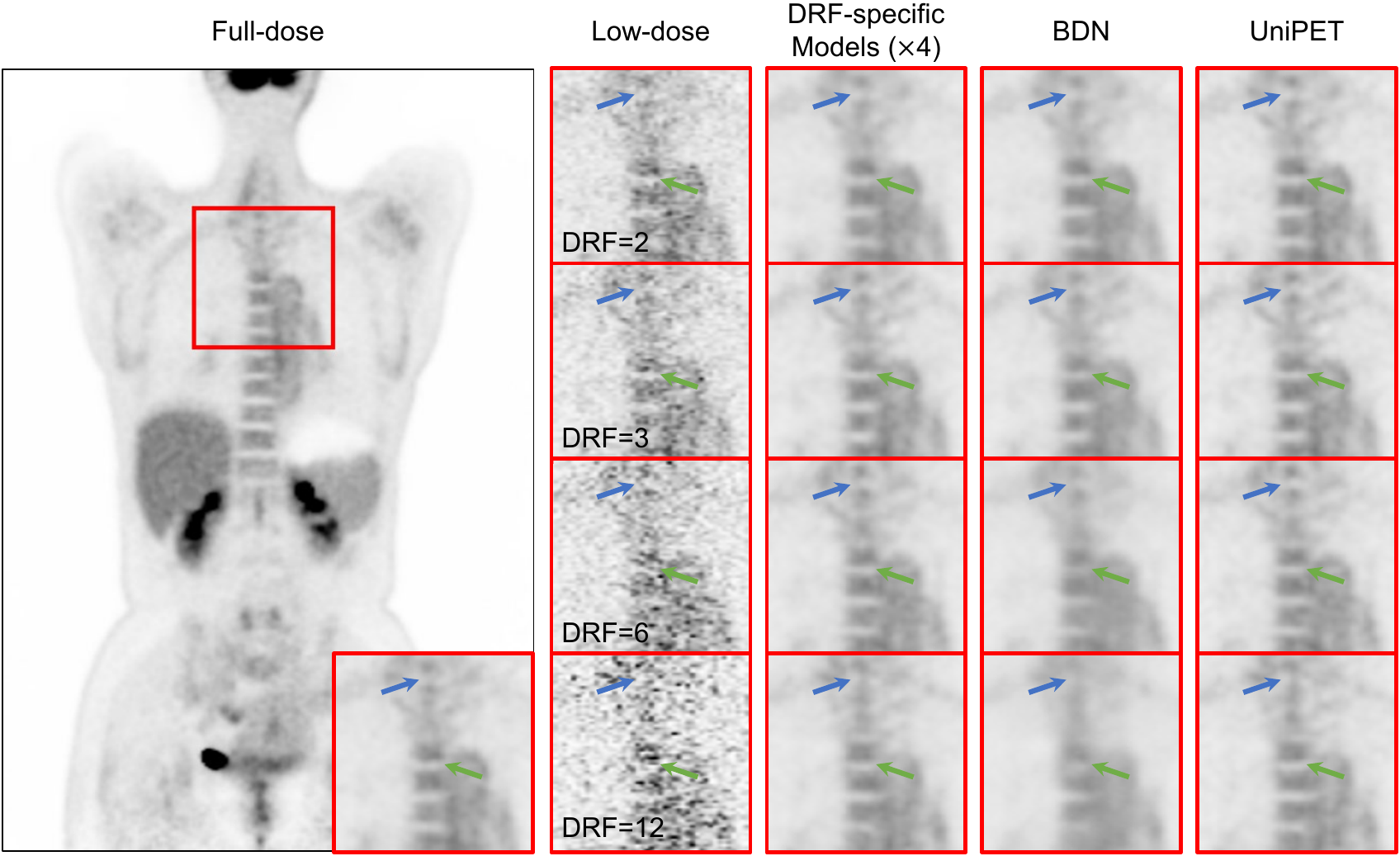}
\caption{Visual comparison between the universal and DRF-specific models on the UPID-Base dataset. Arrows indicate spine regions with notable differences. The proposed universal model UniPET achieves comparable performance to DRF-specific models.} 
\label{fig_single_vs_universal}
\end{figure}   

\begin{figure}[t]
\centering
\includegraphics[width=0.5\textwidth]{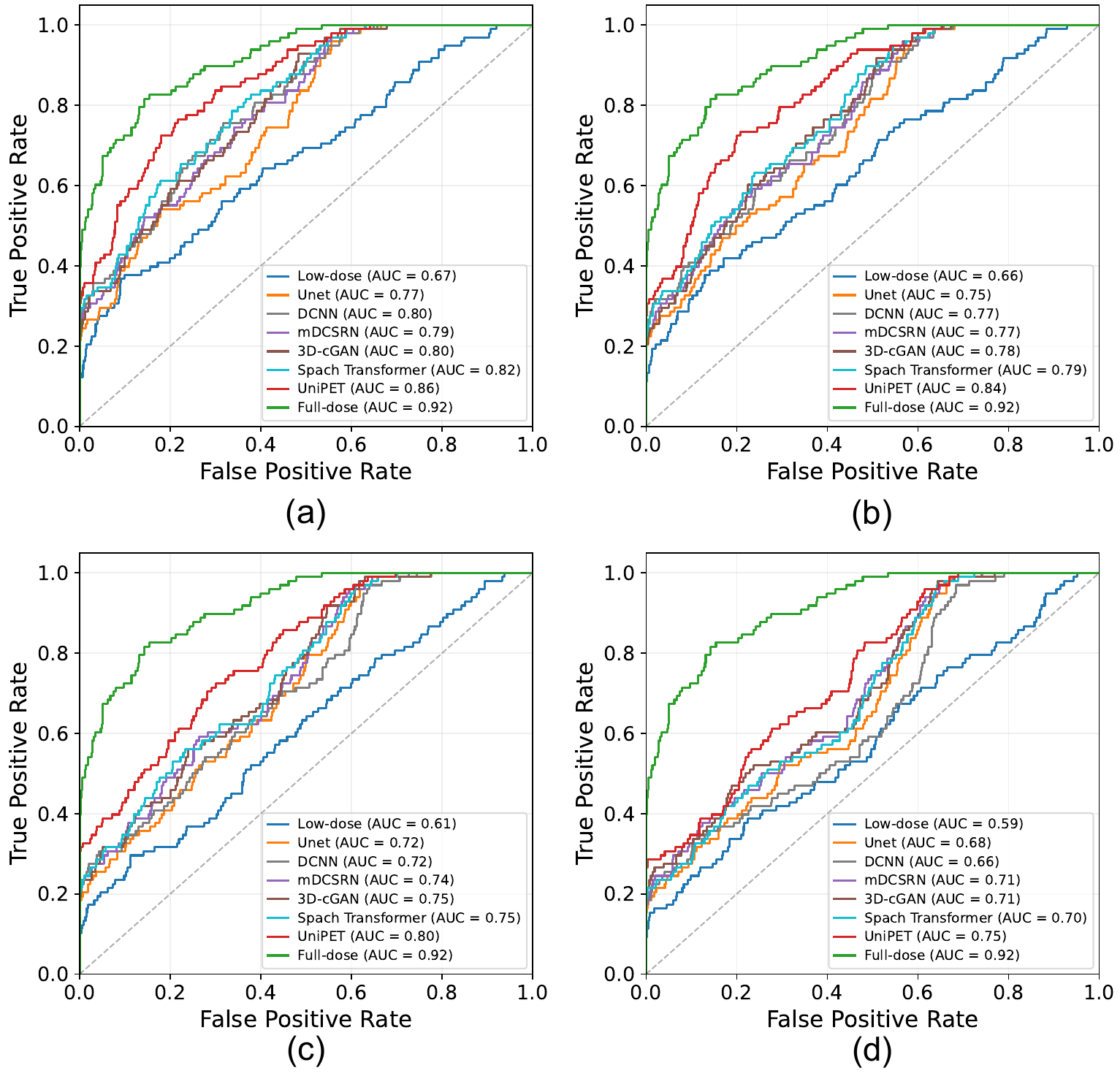}
\caption{ROC curves of different methods on the UPID-Base dataset at different DRFs. (a) DRF=2, (b) DRF=3, (c) DRF=6, and (d) DRF=12.} 
\label{fig_roc}
\end{figure}

\subsection{Comparison methods} 
For comprehensive comparisons, we compare our proposed UniPET with five different 3D medical denoising methods including Unet \citep{schaefferkoetter_Unet_2020}, DCNN \citep{chan_DCNN_2018}, mDCSRN \citep{chen_mDCSRN_2018}, 3D-cGAN \citep{wang_3DcGAN_2018}, and Spach Transformer \citep{jang_spach_transformer_2023}. Except for mDCSRN, which is initially proposed for MRI image super-resolution, all other comparison methods are designed for PET image denoising. Specifically, Unet and DCNN are proposed for universal PET image denoising across different DRFs, while 3D-cGAN and Spach Transformer are proposed for DRF-specific PET image denoising. Among these, Spach Transformer is currently the state-of-the-art model for PET image denoising. These five comparison methods offer a broad spectrum of diversity, as shown in \autoref{table_model_info}. They differ in terms of parameter numbers, model types (CNN and Transformer), architectures (UNet, ResNet, and DenseNet), and the use of GANs. We believe this diversity will facilitate a comprehensive comparison of the models from various perspectives. We train the models for Unet and DCNN following the configurations outlined in their original papers. As for mDCSRN, 3D-cGAN, and Spach Transformer, we re-implement them on our dataset for universal PET image denoising.  

\begin{table*}[!t] 
\caption{Comparisons on the UPID-Base dataset. The best results are highlighted in bold. * Denotes results that are significantly different from the best results, based on a paired t-test with Bonferroni correction.}
\resizebox{\textwidth}{!}{
\begin{tabular}{c|ccccc|ccccc|ccccc}
\hline
\multirow{2}{*}{Method} & \multicolumn{5}{c|}{PSNR↑}                                                         & \multicolumn{5}{c|}{SSIM↑}                                                              & \multicolumn{5}{c}{LPIPS↓}                                                              \\ \cline{2-16} 
                        & DRF=2          & DRF=3          & DRF=6          & DRF=12         & Avg            & DRF=2           & DRF=3           & DRF=6           & DRF=12          & Avg             & DRF=2           & DRF=3           & DRF=6           & DRF=12          & Avg             \\ \hline
Unet                    & 49.48 $^{*}$          & 48.41 $^{*}$          & 46.64 $^{*}$          & 44.58 $^{*}$          & 47.28 $^{*}$          & 0.901 $^{*}$          & 0.893 $^{*}$          & 0.882 $^{*}$          & 0.882 $^{*}$          & 0.889 $^{*}$          & 0.013 $^{*}$          & 0.012 $^{*}$          & 0.016 $^{*}$          & 0.020 $^{*}$          & 0.015 $^{*}$          \\
DCNN                    & 50.91 $^{*}$    & 49.39 $^{*}$    & 47.32 $^{*}$    & 45.06 $^{*}$          & 48.17 $^{*}$    & 0.979 $^{*}$          & 0.972 $^{*}$          & 0.959 $^{*}$    & 0.941 $^{*}$          & 0.963 $^{*}$    & 0.009 $^{*}$          & 0.010 $^{*}$          & 0.016 $^{*}$          & 0.022 $^{*}$          & 0.014 $^{*}$          \\
mDCSRN                  & 49.96 $^{*}$          & 48.71 $^{*}$          & 46.62 $^{*}$          & 44.23 $^{*}$          & 47.38 $^{*}$          & 0.980 $^{*}$    & 0.974 $^{*}$    & 0.959 $^{*}$          & 0.925 $^{*}$          & 0.959 $^{*}$          & 0.014 $^{*}$          & 0.014 $^{*}$          & 0.015 $^{*}$          & 0.022 $^{*}$          & 0.017 $^{*}$          \\
3D-cGAN                 & 49.93 $^{*}$          & 48.72 $^{*}$          & 46.82 $^{*}$          & 44.69 $^{*}$          & 47.54 $^{*}$          & 0.977 $^{*}$          & 0.971 $^{*}$          & 0.957 $^{*}$          & 0.938 $^{*}$          & 0.961 $^{*}$          & 0.010 $^{*}$          & 0.010 $^{*}$          & 0.014 $^{*}$          & 0.017 $^{*}$          & 0.013 $^{*}$          \\
Spach Transformer       & 50.67 $^{*}$          & 49.27 $^{*}$          & 47.29 $^{*}$          & 45.14 $^{*}$    & 48.09 $^{*}$          & 0.976 $^{*}$          & 0.969 $^{*}$          & 0.957 $^{*}$          & 0.945 $^{*}$    & 0.962 $^{*}$          & 0.007 $^{*}$    & 0.008 $^{*}$    & 0.010 $^{*}$    & 0.014 $^{*}$    & 0.010 $^{*}$    \\
UniPET (ours)                   & \textbf{51.54 \textcolor{white}{$^{*}$}} & \textbf{49.66 \textcolor{white}{$^{*}$}} & \textbf{47.55 \textcolor{white}{$^{*}$}} & \textbf{45.45 \textcolor{white}{$^{*}$}} & \textbf{48.55 \textcolor{white}{$^{*}$}} & \textbf{0.982 \textcolor{white}{$^{*}$}} & \textbf{0.975 \textcolor{white}{$^{*}$}} & \textbf{0.963 \textcolor{white}{$^{*}$}} & \textbf{0.947 \textcolor{white}{$^{*}$}} & \textbf{0.967 \textcolor{white}{$^{*}$}} & \textbf{0.004 \textcolor{white}{$^{*}$}} & \textbf{0.006 \textcolor{white}{$^{*}$}} & \textbf{0.008 \textcolor{white}{$^{*}$}} & \textbf{0.011 \textcolor{white}{$^{*}$}} & \textbf{0.007 \textcolor{white}{$^{*}$}} \\ \hline
\end{tabular}} 
\label{table_universal_compare}
\end{table*}

\begin{table*}[!t] 
\caption{Comparisons on the Bern dataset. The best results are highlighted in bold. * Denotes results that are significantly different from the best results, based on a paired t-test with Bonferroni correction.}
\resizebox{\textwidth}{!}{
\begin{tabular}{c|ccccccc|ccccccc|ccccccc}
\hline
\multirow{2}{*}{Method} & \multicolumn{7}{c|}{PSNR↑}                                                                                                                                            & \multicolumn{7}{c|}{SSIM↑}                                                                                                                                                   & \multicolumn{7}{c}{LPIPS↓}                                                                                                                                                   \\ \cline{2-22} 
                        & DRF=2                 & DRF=4                 & DRF=10                & DRF=20                & DRF=50                & DRF=100               & Avg.                  & DRF=2                  & DRF=4                  & DRF=10                 & DRF=20                 & DRF=50                 & DRF=100                & Avg.                   & DRF=2                  & DRF=4                  & DRF=10                 & DRF=20                 & DRF=50                 & DRF=100                & Avg.                   \\ \hline
Unet                    & 51.06 $^{*}$          & 48.37 $^{*}$          & 46.42 $^{*}$          & 45.19 $^{*}$          & 43.19 $^{*}$          & 41.34 $^{*}$          & 45.93 $^{*}$          & 0.985 $^{*}$          & 0.977 $^{*}$          & 0.970 $^{*}$          & 0.964 $^{*}$          & 0.952 $^{*}$          & 0.941 $^{*}$          & 0.965 $^{*}$          & 0.004 $^{*}$          & 0.006 $^{*}$          & 0.008 $^{*}$          & 0.010 $^{*}$          & 0.013 $^{*}$          & 0.016 $^{*}$          & 0.010 $^{*}$          \\
DCNN                    & 51.41 $^{*}$          & 48.70 $^{*}$          & 46.55 $^{*}$          & 45.14 $^{*}$          & 43.09 $^{*}$          & 41.24 $^{*}$          & 46.02 $^{*}$          & 0.947 $^{*}$          & 0.938 $^{*}$          & 0.928 $^{*}$          & 0.923 $^{*}$          & 0.917 $^{*}$          & 0.908 $^{*}$          & 0.927 $^{*}$          & 0.004 $^{*}$          & 0.005 $^{*}$          & 0.007 $^{*}$          & 0.009 $^{*}$          & 0.012 $^{*}$          & 0.016 $^{*}$          & 0.009 $^{*}$          \\
mDCSRN                  & 50.97 $^{*}$          & 48.51 $^{*}$          & 46.32 $^{*}$          & 44.86 $^{*}$          & 42.79 $^{*}$          & 40.94 $^{*}$          & 45.73 $^{*}$          & 0.986 $^{*}$          & 0.979 $^{*}$          & 0.969 $^{*}$          & 0.960 $^{*}$          & 0.948 $^{*}$          & 0.936 $^{*}$          & 0.963 $^{*}$          & 0.005 $^{*}$          & 0.006 $^{*}$          & 0.009 $^{*}$          & 0.012 $^{*}$          & 0.017 $^{*}$          & 0.023 $^{*}$          & 0.012 $^{*}$          \\
3D-cGAN                 & 50.76 $^{*}$          & 48.25 $^{*}$          & 46.09 $^{*}$          & 44.68 $^{*}$          & 42.73 $^{*}$          & 40.97 $^{*}$          & 45.58 $^{*}$          & 0.985 $^{*}$          & 0.978 $^{*}$          & 0.968 $^{*}$          & 0.962 $^{*}$          & 0.951 $^{*}$          & 0.940 $^{*}$          & 0.964 $^{*}$          & 0.003 $^{*}$          & 0.005 $^{*}$          & 0.007 $^{*}$          & 0.009 $^{*}$          & 0.013 $^{*}$          & 0.017 $^{*}$          & 0.009 $^{*}$          \\
Spach Transformer       & 51.40 $^{*}$          & 48.68 $^{*}$          & 46.63 $^{*}$          & 45.23 $^{*}$          & 43.25 $^{*}$          & 41.57 $^{*}$          & 46.13 $^{*}$          & 0.988 \textcolor{white}{$^{*}$}          & 0.979 $^{*}$          & 0.969 $^{*}$          & 0.963 $^{*}$          & 0.954 $^{*}$          & 0.941 $^{*}$          & 0.966 $^{*}$          & 0.003 $^{*}$          & 0.005 $^{*}$          & 0.007 $^{*}$          & 0.009 $^{*}$          & 0.012 $^{*}$          & 0.015 $^{*}$          & 0.009 $^{*}$          \\
UniPET                  & \textbf{51.82 \textcolor{white}{$^{*}$}} & \textbf{48.89 \textcolor{white}{$^{*}$}} & \textbf{46.83 \textcolor{white}{$^{*}$}} & \textbf{45.51 \textcolor{white}{$^{*}$}} & \textbf{43.60 \textcolor{white}{$^{*}$}} & \textbf{41.87 \textcolor{white}{$^{*}$}} & \textbf{46.42 \textcolor{white}{$^{*}$}} & \textbf{0.988 \textcolor{white}{$^{*}$}} & \textbf{0.981 \textcolor{white}{$^{*}$}} & \textbf{0.972 \textcolor{white}{$^{*}$}} & \textbf{0.967 \textcolor{white}{$^{*}$}} & \textbf{0.958 \textcolor{white}{$^{*}$}} & \textbf{0.948 \textcolor{white}{$^{*}$}} & \textbf{0.969 \textcolor{white}{$^{*}$}} & \textbf{0.002 \textcolor{white}{$^{*}$}} & \textbf{0.003 \textcolor{white}{$^{*}$}} & \textbf{0.004 \textcolor{white}{$^{*}$}} & \textbf{0.005 \textcolor{white}{$^{*}$}} & \textbf{0.008 \textcolor{white}{$^{*}$}} & \textbf{0.011 \textcolor{white}{$^{*}$}} & \textbf{0.005 \textcolor{white}{$^{*}$}} \\ \hline
\end{tabular}
}
\label{table_universal_bern_compare}
\end{table*}

\begin{table}[!t] 
\caption{Reader score comparisons on the UPID-Base dataset. 'NR' represents noise reduction, 'SP' represents structure preservation, 'TP' represents texture preservation, and ‘OQ’ represents overall quality. The best results are highlighted in bold. * Denotes results that are significantly different from the best results, based on a paired t-test with Bonferroni correction.}
\centering
\resizebox{0.5\textwidth}{!}{
\begin{tabular}{c|cccc}
\hline
\multirow{2}{*}{Method} & \multicolumn{4}{c}{Reader Score↑}                                                                                                                          \\ \cline{2-5} 
                        & \multicolumn{1}{c|}{NR}      & \multicolumn{1}{c|}{SP} & \multicolumn{1}{c|}{TP} & OQ      \\ \hline
Unet                    & \multicolumn{1}{c|}{3.55 ± 0.43 $^{*}$}          & \multicolumn{1}{c|}{3.15 ± 0.20 $^{*}$}            & \multicolumn{1}{c|}{2.90 ± 0.58 $^{*}$}          & 3.25 ± 0.32 $^{*}$          \\
DCNN                    & \multicolumn{1}{c|}{3.65 ± 0.12 $^{*}$}          & \multicolumn{1}{c|}{3.15 ± 0.12 $^{*}$}            & \multicolumn{1}{c|}{2.10 ± 0.30 $^{*}$}          & 2.45 ± 0.24 $^{*}$          \\
mDCSRN                  & \multicolumn{1}{c|}{3.30 ± 0.29 $^{*}$}          & \multicolumn{1}{c|}{2.70 ± 0.24 $^{*}$}            & \multicolumn{1}{c|}{3.20 ± 0.48 $^{*}$}          & 2.65 ± 0.54 $^{*}$          \\
3D-cGAN                 & \multicolumn{1}{c|}{4.00 ± 0.27 $^{*}$}          & \multicolumn{1}{c|}{3.55 ± 0.19 $^{*}$}            & \multicolumn{1}{c|}{2.95 ± 0.33 $^{*}$}          & 3.15 ± 0.12 $^{*}$          \\
Spach Transformer       & \multicolumn{1}{c|}{4.20 ± 0.10 $^{*}$}   & \multicolumn{1}{c|}{4.20 ± 0.19 $^{*}$}      & \multicolumn{1}{c|}{3.30 ± 0.48 $^{*}$}    &  3.85 ± 0.20 $^{*}$    \\
UniPET (ours)                   & \multicolumn{1}{c|}{\textbf{4.60 ± 0.20 \textcolor{white}{$^{*}$}}} & \multicolumn{1}{c|}{\textbf{4.40 ± 0.12 \textcolor{white}{$^{*}$}}}   & \multicolumn{1}{c|}{\textbf{4.35 ± 0.30 \textcolor{white}{$^{*}$}}} & \textbf{4.60 ± 0.12 \textcolor{white}{$^{*}$}} \\ \hline
\end{tabular}
}
\label{table_reader_study}
\end{table}

\begin{table}[!t]

\caption{Clinical comparisons on the UPID-Base dataset. The best results are highlighted in bold. * Denotes results that are significantly different from the best results, based on a paired t-test with Bonferroni correction.}
\centering
\resizebox{0.5\textwidth}{!}{
\begin{tabular}{c|ccc}
\hline
\multirow{2}{*}{Method} & \multicolumn{3}{c}{MAE↓}                                                                                                 \\ \cline{2-4} 
                        & \multicolumn{1}{c|}{Blood Pool}               & \multicolumn{1}{c|}{Liver}                    & Lesion                   \\ \hline
Unet                    & \multicolumn{1}{c|}{0.113 ± 0.026 $^{*}$}          & \multicolumn{1}{c|}{0.104 ± 0.025 $^{*}$}          & 0.195 ± 0.036 $^{*}$           \\
DCNN                    & \multicolumn{1}{c|}{0.111 ± 0.028 $^{*}$}          & \multicolumn{1}{c|}{0.107 ± 0.022 $^{*}$}          & 0.178 ± 0.048 $^{*}$    \\
mDCSRN                  & \multicolumn{1}{c|}{0.112 ± 0.030 $^{*}$}          & \multicolumn{1}{c|}{0.106 ± 0.031 $^{*}$}          & 0.198 ± 0.049 $^{*}$          \\
3D-cGAN                 & \multicolumn{1}{c|}{0.109 ± 0.027 $^{*}$}          & \multicolumn{1}{c|}{0.100 ± 0.024 $^{*}$}          & 0.192 ± 0.045 $^{*}$          \\
Spach Transformer       & \multicolumn{1}{c|}{0.104 ± 0.025 $^{*}$}    & \multicolumn{1}{c|}{0.099 ± 0.023 \textcolor{white}{$^{*}$}}    & 0.189 ± 0.043 $^{*}$          \\
UniPET (ours)                   & \multicolumn{1}{c|}{\textbf{0.098 ± 0.027 \textcolor{white}{$^{*}$}}} & \multicolumn{1}{c|}{\textbf{0.099 ± 0.023 \textcolor{white}{$^{*}$}}} & \textbf{0.165 ± 0.052 \textcolor{white}{$^{*}$}} \\ \hline
\end{tabular}
} 
\label{table_clinical}
\end{table}

\begin{table}[!t] 
\caption{Lesion detection performance comparison on the UPID-Base dataset. The best results are highlighted in bold.}
\centering
\resizebox{0.5\textwidth}{!}{
\begin{tabular}{c|cccccccccccc}
\hline
\multirow{2}{*}{Method} & \multicolumn{12}{c}{F1-Score↑}                                                                                                                              \\ \cline{2-13} 
                        & \multicolumn{3}{c|}{DRF=2}               & \multicolumn{3}{c|}{DRF=3}               & \multicolumn{3}{c|}{DRF=6}               & \multicolumn{3}{c}{DRF=12} \\ \hline
Low-dose                &  & 0.94          & \multicolumn{1}{c|}{} &  & 0.84          & \multicolumn{1}{c|}{} &  & 0.68          & \multicolumn{1}{c|}{} &     & 0.58            &    \\
Unet                    &  & 0.98          & \multicolumn{1}{c|}{} &  & 0.94          & \multicolumn{1}{c|}{} &  & 0.89          & \multicolumn{1}{c|}{} &     & 0.79            &    \\
DCNN                    &  & 0.99          & \multicolumn{1}{c|}{} &  & 0.96          & \multicolumn{1}{c|}{} &  & 0.91          & \multicolumn{1}{c|}{} &     & 0.81            &    \\
mDCSRN                  &  & 0.99          & \multicolumn{1}{c|}{} &  & 0.96          & \multicolumn{1}{c|}{} &  & 0.88          & \multicolumn{1}{c|}{} &     & 0.79            &    \\
3D-cGAN                 &  & 0.98          & \multicolumn{1}{c|}{} &  & 0.96          & \multicolumn{1}{c|}{} &  & 0.92          & \multicolumn{1}{c|}{} &     & 0.81            &    \\
Spach Transformer       &  & 0.99          & \multicolumn{1}{c|}{} &  & 0.97          & \multicolumn{1}{c|}{} &  & 0.95          & \multicolumn{1}{c|}{} &     & 0.84            &    \\
UniPET (ours)           &  & \textbf{1.00} & \multicolumn{1}{c|}{} &  & \textbf{0.99} & \multicolumn{1}{c|}{} &  & \textbf{0.97} & \multicolumn{1}{c|}{} &     & \textbf{0.93}   &    \\ \hline
\end{tabular}} 
\label{table_detection_rate}
\end{table}

\section{Experimental results} 

\subsection{Comparisons between universal and DRF-specific models} 
We compare the performance between the universal model and DRF-specific models on the UPID-Base dataset. For a fair comparison, we augment the low-dose data used for training the DRF-specific models by performing four different random list mode decimations for each DRF, thereby ensuring an equivalent training data volume for both the DRF-specific and universal models. In addition, we implement BDN without SAN and RALS as our baseline universal model. The experimental outcomes are depicted in \autoref{table_universal_vs_specific}. The DRF-specific model performs well on the DRF it was trained on. However, its performance drops significantly on other DRF datasets, leading to an overall lower performance across all DRFs. BDN achieves relatively good overall performance in PSNR. Nonetheless, due to the \textit{style elimination issue}, it shows weaker performance on specific DRFs compared to the DRF-specific models, particularly in metrics reflecting style (LPIPS). Our proposed UniPET significantly mitigates the \textit{style elimination issue} and bolsters BDN by more effectively preserving styles. UniPET outperforms BDN across all DRF datasets, delivering average gains of 0.47 dB in PSNR, 0.005 in SSIM, and a 0.009 reduction in LPIPS. Moreover, for each individual DRF, UniPET achieves results comparable to those of the DRF-specific models.

\autoref{fig_single_vs_universal} presents an example of a visual comparison of the spine. The visualization results align well with the quantitative results in \autoref{table_universal_vs_specific}.
DRF-specific models excel at specific DRFs and effectively recover the structures and edges of the spine. However, the vanilla universal model BDN encounters a significant over-smoothing effect attributed to the \textit{style elimination issue}. Our proposed UniPET consistently performs well across DRFs, achieving results comparable to DRF-specific models.

\begin{figure*}[t]
\centering
\includegraphics[width=0.9\textwidth]{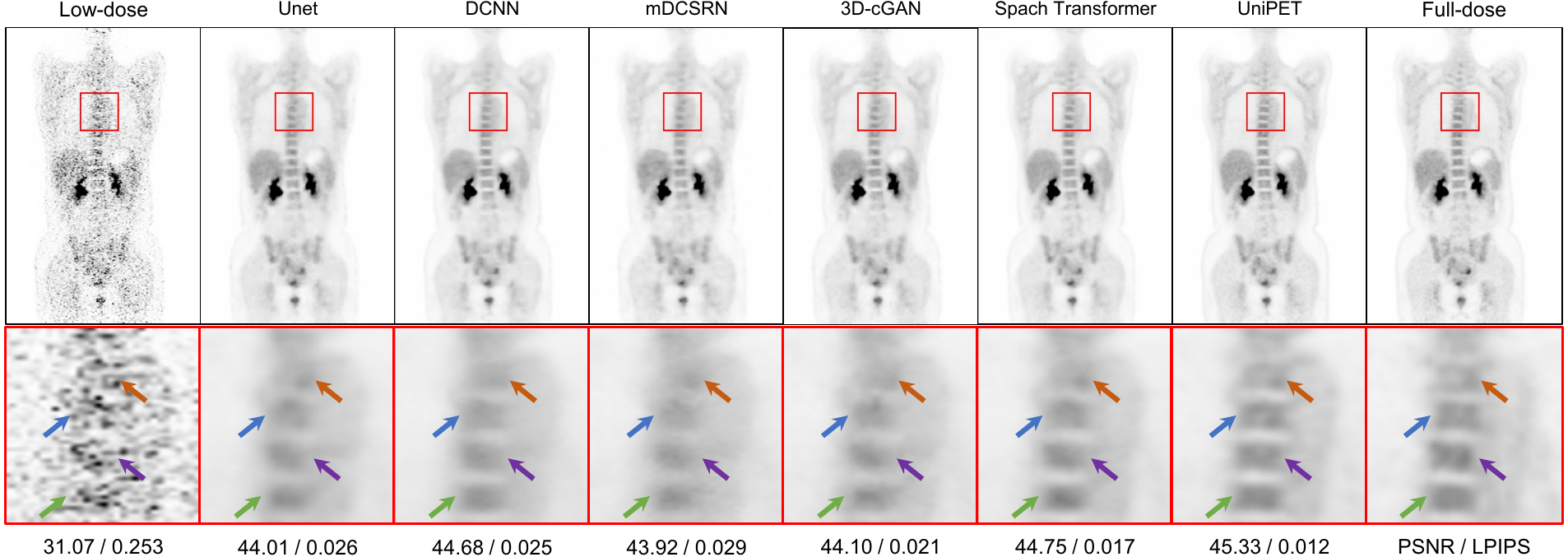}
\caption{Visual comparison of different methods on the UPID-Base dataset. The input low-dose PET image corresponds to DRF = 12. The zoomed-in rectangular region is recommended for better visualization. Arrows indicate spine regions with notable differences.} 
\label{fig_method_comparison_spine}
\end{figure*}

\begin{figure*}[t]
\centering
\includegraphics[width=0.9\textwidth]{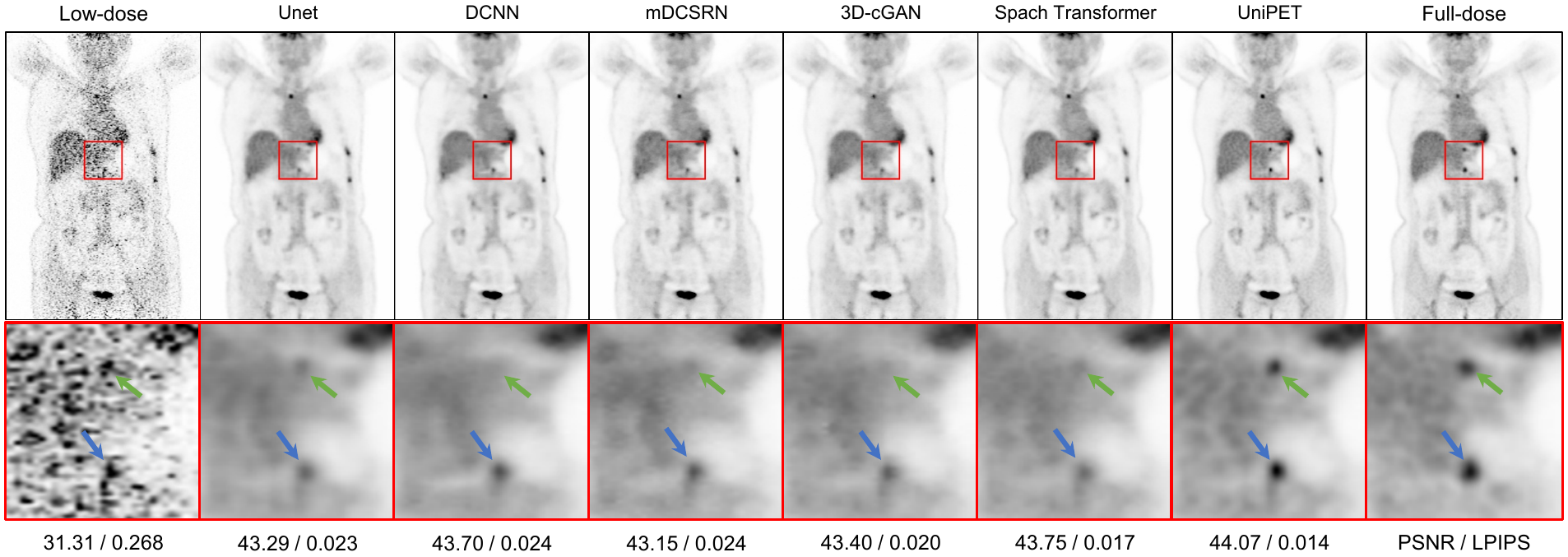}
\caption{Visual comparison of different methods on the UPID-Base dataset. The input low-dose PET image corresponds to DRF = 12. The zoomed-in rectangular region is recommended for better visualization. Arrows indicate lesion regions with notable differences.} 
\label{fig_method_comparison_lesion}
\end{figure*}

\begin{figure*}[t]
\centering
\includegraphics[width=0.9\textwidth]{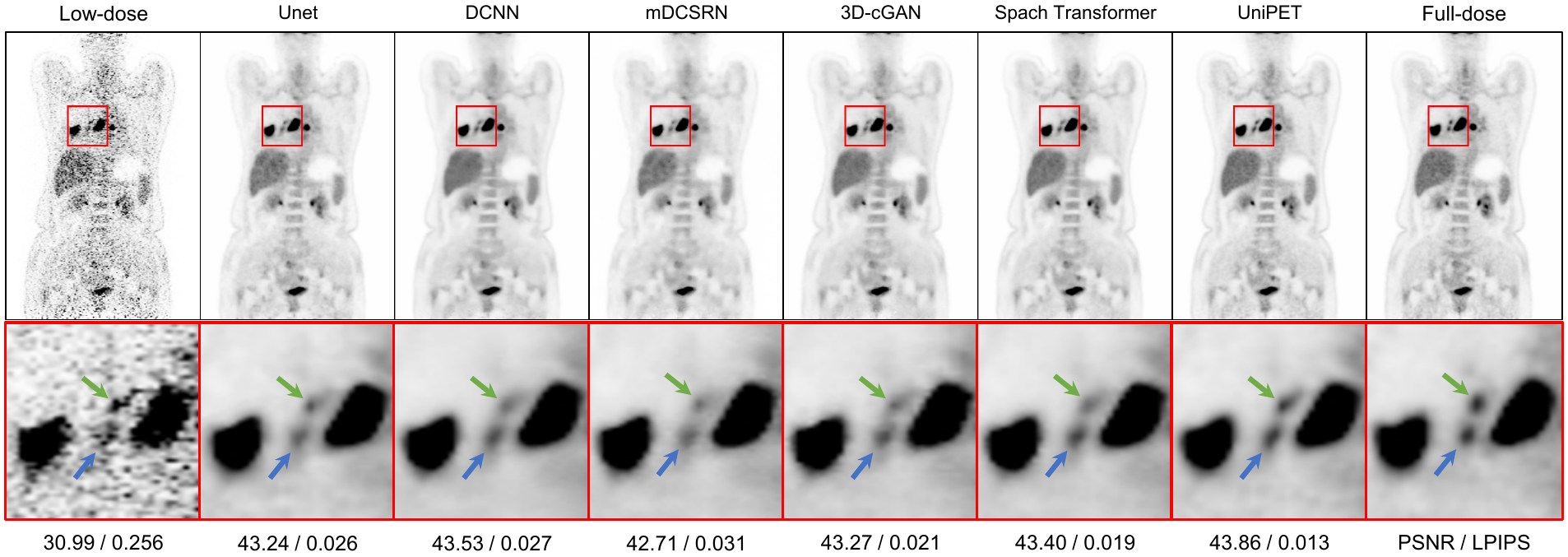}
\caption{Visual comparison of different methods on the UPID-Base dataset. The input low-dose PET image corresponds to DRF=12. The zoomed-in rectangular region is recommended for better visualization. Arrows indicate lesion regions with notable differences. }
\label{fig_method_comparison}
\end{figure*}  

\begin{figure*}[t]
\centering
\includegraphics[width=0.9\textwidth]{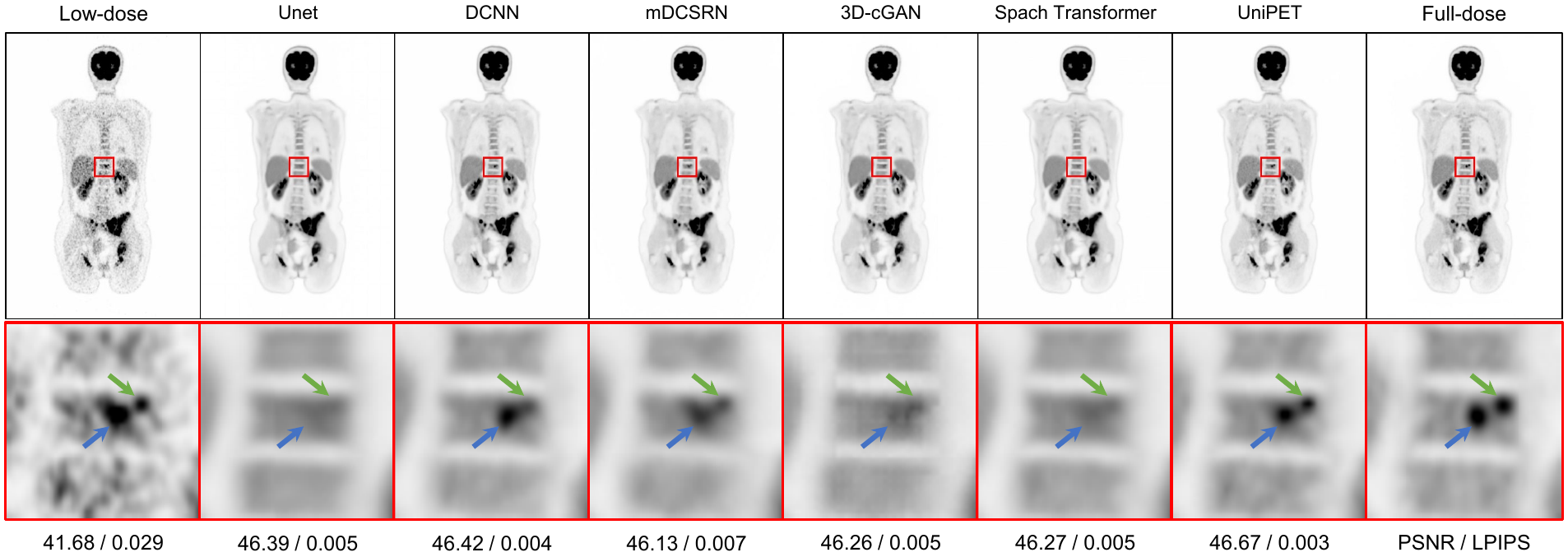}
\caption{Visual comparison of different methods on the Bern dataset. The input low-dose PET image corresponds to DRF = 20. The zoomed-in rectangular region is recommended for better visualization. Arrows indicate lesion regions with notable differences.} 
\label{fig_method_comparison_Siemens_lesion}
\end{figure*}  

\begin{figure*}[t]
\centering
\includegraphics[width=0.9\textwidth]{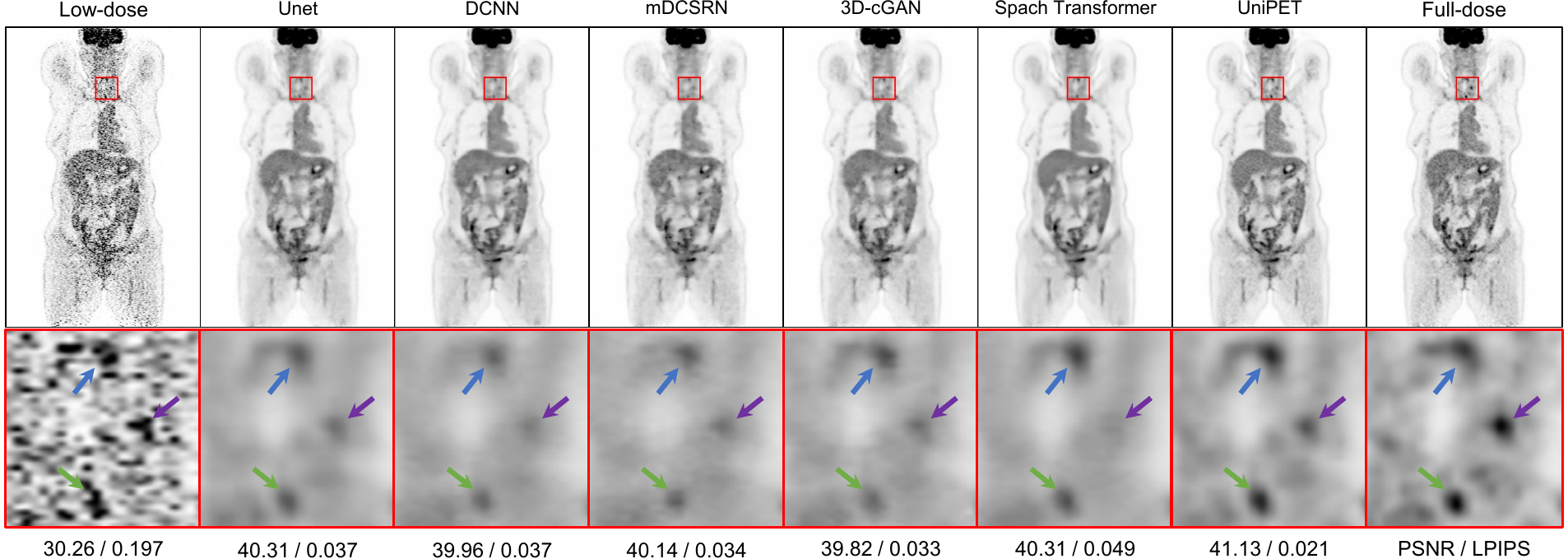}
\caption{Visual comparison of different methods on the UPID-OOD-Center dataset. The input low-dose PET image corresponds to DRF=4. The zoomed-in rectangular region is recommended for better visualization. Arrows indicate lesion regions with notable differences.} 
\label{fig_ood_center}
\end{figure*}

\subsection{Comparisons on the UPID-Base and Bern datasets} 
We compare our proposed UniPET with other state-of-the-art denoising methods on UPID-Base and Bern datasets. As depicted in \autoref{table_universal_compare}, \autoref{table_universal_bern_compare}, \autoref{table_reader_study}, \autoref{table_clinical}, and \autoref{table_detection_rate},  UniPET showcases superiority over all comparison methods in quantity, perception, and clinical evaluation. In \autoref{table_universal_compare} and \autoref{table_universal_bern_compare}, UniPET demonstrates a balanced performance between quantitative and perceptual metrics, particularly excelling in the perceptual metric LPIPS compared to other comparative methods. The reader score comparison results in \autoref{table_reader_study} consistently depict UniPET's outstanding performance in noise reduction, structure preservation, texture preservation, and overall image quality. This suggests that the synthesized images produced by UniPET are more perceptually appealing to radiologists. \autoref{table_clinical} shows the SUV errors of different methods across three ROIs. According to the paper \citep{scheuermann2009PET_qualification}, an average SUV error of 10\% is clinically acceptable in PET images. In the UPID‑Base testing set, the mean SUV values are 1.862 in the blood pool, 1.906 in the liver, and 3.447 in the lesion. Accordingly, the corresponding clinically acceptable thresholds of SUV error are 0.1862 for the blood pool, 0.1906 for the liver, and 0.3447 for the lesion. As \autoref{table_clinical} shows, every method’s SUV error in all three ROIs falls below these thresholds, and is therefore clinically acceptable on average. Among them, our proposed UniPET achieves the lowest overall SUV error and best preserves uptake values in all three regions. \autoref{table_detection_rate} presents the averaged F1-score of the lesion detection task performed by three radiologists on PET images restored by various methods. As shown, low‑dose PET images exhibit poor lesion detectability: for very low‑dose images (DRF = 12), the average F1‑score is only 0.58. This low performance stems from significant noise, which can both obscure small lesions and produce false positives. By comparison, all deep‑learning–based denoising methods substantially improve lesion detection performance by effectively reducing noise while preserving lesion structures. Our proposed UniPET consistently achieves the best lesion detection performance across all DRFs (F1-score $>$ 0.9), indicating that it most effectively recovers lesion regions identifiable by radiologists. In \autoref{fig_roc}, the ROC curves for all low-dose images lie well below that of the full-dose reference, with the smallest areas under the curves (AUCs) observed at 0.67 (DRF=2), 0.66 (DRF=3), 0.61 (DRF=6), and 0.59 (DRF=12), indicating markedly poorer lesion detectability. All of the deep learning–based denoising methods improve lesion detectability in low-dose PET images. Among them, the ROC curve of UniPET consistently lies closest to that of the full-dose data, achieving the highest AUCs across all DRFs—0.86 (DRF=2), 0.84 (DRF=3), 0.80 (DRF=6), and 0.75 (DRF=12)—demonstrating that UniPET produces images whose overall quality and lesion detectability most closely match those of full-dose images. 

The visual comparison results are shown in \autoref{fig_method_comparison_spine}, \autoref{fig_method_comparison_lesion}, \autoref{fig_method_comparison}, and \autoref{fig_method_comparison_Siemens_lesion}.  The visualization results align well with the reader scores in \autoref{table_reader_study}. While all methods achieve acceptable denoising performance, our proposed UniPET excels in recovering both anatomical structures and fine textures, including spatial patterns and lesions. Specifically, \autoref{fig_method_comparison_spine} shows that UniPET more effectively recovers the spine structures, while the comparison methods exhibit varying degrees of over-smoothing. \autoref{fig_method_comparison_lesion}, \autoref{fig_method_comparison}, and \autoref{fig_method_comparison_Siemens_lesion} present visual comparisons of lesion regions. In \autoref{fig_method_comparison_lesion}, the lesions in the low-dose image are partially obscured by surrounding noise. While all deep learning methods effectively reduce background noise, several (e.g., DCNN, mDCSRN, D-cGAN, and Spach Transformer) appear to excessively smooth the image, leading to partial or complete removal of the upper lesion. In contrast, the proposed UniPET method retains both lesions with improved detectability. In \autoref{fig_method_comparison}, although all methods are able to show the presence of lesions, UniPET achieves the best recovery of lesion contrast, offering clearer and more defined lesion boundaries. In \autoref{fig_method_comparison_Siemens_lesion}, despite the low-dose image already exhibiting good lesion detectability, some methods (e.g., Unet, 3D-cGAN, and Spach Transformer) mistakenly remove lesions and thus impair detectability. These qualitative observations align with the quantitative detection performance reported in \autoref{table_detection_rate}, where UniPET achieves the best lesion detectability (F1-score $>$ 0.9) by radiologists.

\begin{figure*}[t]
\centering
\includegraphics[width=0.75\textwidth]{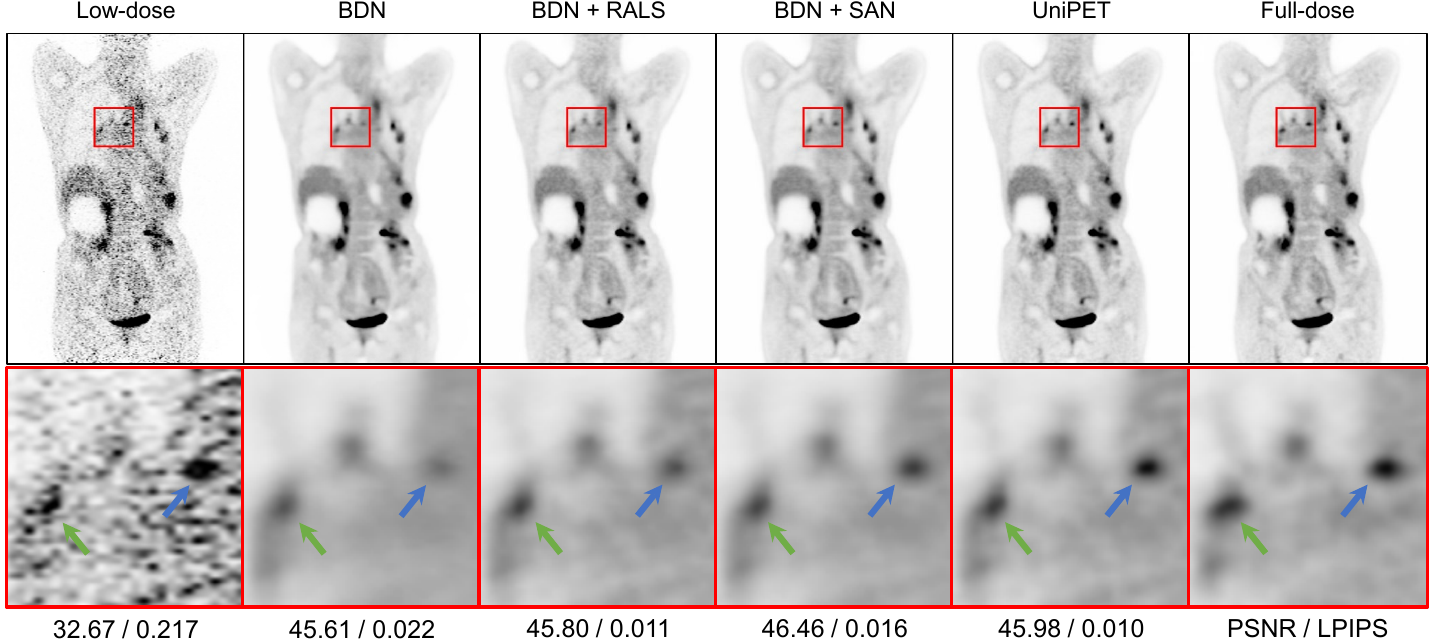}
\caption{Visual comparison for component analysis on the UPID-Base dataset. The input low-dose PET image corresponds to DRF=12. The zoomed-in rectangular region is recommended for better visualization. Arrows indicate lesion regions with notable differences. }
\label{fig_component_analysis}
\end{figure*}

\begin{table}[t] 

\caption{Comparisons on the UPID-OOD-DRF dataset. The best results are highlighted in bold. * Denotes results that are significantly different from the best results, based on a paired t-test with Bonferroni correction.}
\centering
\resizebox{0.5\textwidth}{!}{
\begin{tabular}{c|c|c|c}
\hline
Method            & PSNR↑                   & SSIM↑                    & LPIPS↓                   \\ \hline
Unet              & 46.78 ± 1.79 $^{*}$          & 0.887 ± 0.012 $^{*}$          & 0.015 ± 0.002 $^{*}$          \\
DCNN              & 47.58 ± 2.08 $^{*}$    & 0.962 ± 0.015 $^{*}$    & 0.013 ± 0.004 $^{*}$          \\
mDCSRN            & 47.09 ± 1.71 $^{*}$          & 0.962 ± 0.018 $^{*}$          & 0.016 ± 0.002 $^{*}$          \\
3D-cGAN           & 47.04 ± 1.79 $^{*}$          & 0.961 ± 0.014 $^{*}$           & 0.013 ± 0.002 $^{*}$          \\
Spach Transformer & 47.49 ± 1.92 $^{*}$          & 0.961 ± 0.012 $^{*}$          & 0.009 ± 0.002 $^{*}$    \\
UniPET (ours)             & \textbf{47.93 ± 2.28 \textcolor{white}{$^{*}$}} & \textbf{0.967 ± 0.013 \textcolor{white}{$^{*}$}} & \textbf{0.007 ± 0.003 \textcolor{white}{$^{*}$}} \\ \hline
\end{tabular}
} 
\label{table_ood_drf}
\end{table}

\begin{table}[t] 

\caption{Comparisons on the UPID-OOD-Center dataset. The best results are highlighted in bold. * Denotes results that are significantly different from the best results, based on a paired t-test with Bonferroni correction.}
\centering
\resizebox{0.5\textwidth}{!}{
\begin{tabular}{c|c|c|c}
\hline
Method            & PSNR↑                 & SSIM↑                    & LPIPS↓                   \\ \hline
Unet              & 42.54 ± 1.81 $^{*}$    & 0.855 ± 0.010 $^{*}$          & 0.039 ± 0.001 $^{*}$          \\
DCNN              & 42.31 ± 1.79 $^{*}$          & 0.948 ± 0.012 $^{*}$          & 0.037 ± 0.003 $^{*}$          \\
mDCSRN            & 42.51 ± 1.68 $^{*}$          & 0.956 ± 0.016 $^{*}$    & 0.032 ± 0.003 $^{*}$    \\
3D-cGAN           & 42.03 ± 1.62 $^{*}$          & 0.950 ± 0.013 $^{*}$          & 0.036 ± 0.001 $^{*}$          \\
Spach Transformer & 42.50 ± 1.89 $^{*}$          & 0.928 ± 0.010 $^{*}$          & 0.050 ± 0.005 $^{*}$          \\
UniPET (ours)     & \textbf{43.61 ± 1.89 \textcolor{white}{$^{*}$}} & \textbf{0.963 ± 0.013 \textcolor{white}{$^{*}$}} & \textbf{0.021 ± 0.003 \textcolor{white}{$^{*}$}} \\ \hline
\end{tabular}

} 
\label{table_ood_center}
\end{table}

\subsection{Comparisons on the UPID-OOD-DRF and UPID-OOD-Center datasets} 
To further examine the model's generalizability on out-of-distribution (OOD) data, we test the performance of different methods on data from unknown DRFs and unknown centers. On the UPID-OOD-DRF dataset, we assess the model's performance on low-dose data with unknown DRFs, as shown in \autoref{table_ood_drf}. UniPET outperforms all comparison methods, surpassing the second‐best Spach Transformer by 0.44 dB in PSNR, 0.006 in SSIM, and 0.002 in LPIPS. These improvements demonstrate UniPET’s strong generalizability across varying DRFs. On the UPID-OOD-Center dataset, we assess the model's performance on low-dose data from an unknown center with distinct imaging protocols, as shown in \autoref{table_ood_center}. Due to the significant domain shift between centers, different methods exhibit a substantial performance drop on the UPID-OOD-Center dataset. Our proposed UniPET demonstrates the strongest generalizability, surpassing the second-best mDCSRN by 1.10 dB in PSNR, 0.007 in SSIM, and 0.011 in LPIPS.

\autoref{fig_ood_center} shows a visual comparison of lesion regions from the UPID-OOD-Center dataset. When processing images from an unknown center, most comparative methods exhibit significant over-smoothing effects due to their inability to handle domain shift. Notably, the Spach Transformer completely smooths out a small lesion. Our proposed UniPET best preserves the contrast and detectability of small lesions.

\begin{table}[t]
\caption{Lesion detection performance of SAN and RALS. The best results are highlighted in bold.}
\centering
\resizebox{0.5\textwidth}{!}{
\begin{tabular}{cc|cccccccccccc}
\hline
\multirow{2}{*}{SAN} & \multirow{2}{*}{RALS} & \multicolumn{12}{c}{F1-Score↑}                                                                                                                                                                             \\ \cline{3-14} 
                     &                       & \multicolumn{3}{c|}{DRF=2}                        & \multicolumn{3}{c|}{DRF=3}                                 & \multicolumn{3}{c|}{DRF=6}                                 & \multicolumn{3}{c}{DRF=12}   \\ \hline
                     &                       &  & 0.99          & \multicolumn{1}{c|}{}          &           & 0.95          & \multicolumn{1}{c|}{}          &           & 0.92          & \multicolumn{1}{c|}{}          &           & 0.83          &  \\
                     & $\checkmark$          &  & 0.99          & \multicolumn{1}{c|}{}          &           & 0.97          & \multicolumn{1}{c|}{}          &           & 0.94          & \multicolumn{1}{c|}{}          &           & 0.85          &  \\
$\checkmark$         &                       &  & 0.99          & \multicolumn{1}{c|}{}          &           & 0.97          & \multicolumn{1}{c|}{}          &           & 0.95          & \multicolumn{1}{c|}{}          &           & 0.88          &  \\
$\checkmark$         & $\checkmark$          &  & \textbf{1.00} & \multicolumn{1}{c|}{\textbf{}} & \textbf{} & \textbf{0.99} & \multicolumn{1}{c|}{\textbf{}} & \textbf{} & \textbf{0.97} & \multicolumn{1}{c|}{\textbf{}} & \textbf{} & \textbf{0.93} &  \\ \hline
\end{tabular}
} 
\label{table_component_analysis_lesion_detection}
\end{table} 

\begin{table}[t]
\caption{AUC from ROC analysis of SAN and RALS. The best results are highlighted in bold.}
\centering
\resizebox{0.5\textwidth}{!}{
\begin{tabular}{cc|cccccccccccc}
\hline
\multirow{2}{*}{SAN} & \multirow{2}{*}{RALS} & \multicolumn{12}{c}{AUC ↑}                                                                                                                                                                                 \\ \cline{3-14} 
                     &                       & \multicolumn{3}{c|}{DRF=2}                        & \multicolumn{3}{c|}{DRF=3}                                 & \multicolumn{3}{c|}{DRF=6}                                 & \multicolumn{3}{c}{DRF=12}   \\ \hline
                     &                       &  & 0.81          & \multicolumn{1}{c|}{}          &           & 0.78          & \multicolumn{1}{c|}{}          &           & 0.73          & \multicolumn{1}{c|}{}          &           & 0.68          &  \\
                     & $\checkmark$          &  & 0.83          & \multicolumn{1}{c|}{}          &           & 0.80          & \multicolumn{1}{c|}{}          &           & 0.74          & \multicolumn{1}{c|}{}          &           & 0.71          &  \\
$\checkmark$         &                       &  & 0.84          & \multicolumn{1}{c|}{}          &           & 0.81          & \multicolumn{1}{c|}{}          &           & 0.76          & \multicolumn{1}{c|}{}          &           & 0.72          &  \\
$\checkmark$         & $\checkmark$          &  & \textbf{0.86} & \multicolumn{1}{c|}{\textbf{}} & \textbf{} & \textbf{0.84} & \multicolumn{1}{c|}{\textbf{}} & \textbf{} & \textbf{0.80} & \multicolumn{1}{c|}{\textbf{}} & \textbf{} & \textbf{0.75} &  \\ \hline
\end{tabular}
} 
\label{table_component_analysis_auc}
\end{table}

\subsection{Ablation study} 
To investigate the significance of individual components in UniPET, we conduct ablation studies on the UPID-Base dataset specifically focusing on two pivotal components: the style alignment network (SAN) and the region-aware learning strategy (RALS).

\textbf{(1) Effect of SAN and RALS.} The component analysis in \autoref{table_component_analysis} delineates the significance of SAN and RALS. Both SAN and RALS play pivotal roles in enhancing UniPET's performance. SAN substantially improves performance across all four evaluation metrics by aligning image styles. RALS primarily enhances style recovery (LPIPS) and lesion restoration (MAE) by directing the model’s attention to the stylized region. The combination of SAN and RALS produces the best lesion recovery performance, yielding a 0.024 decrease in lesion-region MAE compared with the baseline. \autoref{table_component_analysis_lesion_detection} reports lesion detection performance assessed by radiologists. As shown, both SAN and RALS improve lesion detectability, particularly at higher DRFs. At DRF = 12, RALS alone increases F1-score by 0.02 relative to the baseline, SAN alone by 0.05, and the combined method by 0.10. \autoref{table_component_analysis_auc} reports AUCs from ROC analysis of the lesion classification model. Compared with the baseline, both SAN and RALS consistently improve AUC across DRFs, and their combination yields the largest gain: 0.05 at DRF = 2, 0.06 at DRF = 3, and 0.07 at DRF = 6 and 12. These results indicate that both SAN and RALS can enhance lesion detectability. The example of visual comparison depicted in \autoref{fig_component_analysis} clearly illustrates the effectiveness of both SAN and RALS in enhancing the quality of low-dose PET images, particularly in recovering small lesions.  The MAE measured within lesions of \autoref{fig_component_analysis} are: low-dose (0.337), BDN (0.197), BDN+RALS (0.190), BDN+SAN (0.185), and UniPET (0.177). Compared with BDN, BDN+RALS reduces lesion MAE by 0.007, BDN+SAN by 0.012, and UniPET by 0.020. These quantitative improvements are consistent with the visual results in \autoref{fig_component_analysis}, where the use of RALS and SAN yields higher lesion contrast and sharper lesion boundaries. 

\begin{table}[!t]

\caption{Component analysis of SAN and RALS. The best results are highlighted in bold, while the second-best results are underlined. * Denotes results that are significantly different from the best results, based on a paired t-test with Bonferroni correction.}
\centering
\resizebox{0.5\textwidth}{!}{
\begin{tabular}{cc|c|c|c|c}
\hline
\multicolumn{1}{c}{SAN} & \multicolumn{1}{c|}{RALS} & PSNR↑                                                                & SSIM↑                                                                   & LPIPS↓                                                                  & MAE (Lesion)↓                                                           \\ \hline
                        &                           & 48.08 ± 2.11 $^{*}$                                                  & 0.962 ± 0.019 $^{*}$                                                    & 0.016 ± 0.005 $^{*}$                                                    & 0.189 ± 0.042 $^{*}$                                                    \\
                        & $\checkmark$              & 47.85 ± 2.00 $^{*}$                                                  & 0.965 ± 0.017 $^{*}$                                                    & {\ul 0.009 ± 0.003 $^{*}$}                                              & 0.173 ± 0.047 $^{*}$                                                    \\
$\checkmark$            &                           & \textbf{48.98 ± 2.26 \textcolor{white}{$^{*}$}} & \textbf{0.968 ± 0.014 \textcolor{white}{$^{*}$}} & 0.011 ± 0.005 $^{*}$                                                    & {\ul 0.172 ± 0.047 $^{*}$}                                              \\
$\checkmark$            & $\checkmark$              & {\ul 48.55 ± 2.28 $^{*}$}                                            & {\ul 0.967 ± 0.014 $^{*}$}                                              & \textbf{0.007 ± 0.003 \textcolor{white}{$^{*}$}} & \textbf{0.165 ± 0.052 \textcolor{white}{$^{*}$}} \\ \hline
\end{tabular}
} 
\label{table_component_analysis}
\end{table} 

\begin{table}[!t] 
\caption{Performance of the framework on various denoising networks with different network architectures. The symbol '$\dag$' denotes applying SAN and RALS to the corresponding denoising network. The best results within each architecture are highlighted in bold. * Denotes results that are significantly different from the best results, based on a paired t-test with Bonferroni correction.}
\centering
\resizebox{0.5\textwidth}{!}{
\begin{tabular}{c|c|c|c|c|c}
\hline
Architecture              & Method                                               & PSNR↑                                                                & SSIM↑                                                                   & LPIPS↓                                                                  & MAE (Lesion)↓                                                           \\ \hline
\multirow{2}{*}{ResNet}   & BDN \textcolor{white}{$\dag$}     & 48.08 ± 2.11 $^{*}$                                                  & 0.962 ± 0.019 $^{*}$                                                    & 0.016 ± 0.005 $^{*}$                                                     & 0.189 ± 0.042 $^{*}$                                                     \\
                          & BDN $\dag$                                           & \textbf{48.55 ± 2.28 \textcolor{white}{$^{*}$}} & \textbf{0.967 ± 0.014 \textcolor{white}{$^{*}$}} & \textbf{0.007 ± 0.003 \textcolor{white}{$^{*}$}} & \textbf{0.165 ± 0.052 \textcolor{white}{$^{*}$}} \\ \hline
\multirow{2}{*}{DenseNet} & mDCSRN \textcolor{white}{$\dag$}  & 47.38 ± 1.98 $^{*}$                                                  & 0.959 ± 0.022 $^{*}$                                                    & 0.017 ± 0.004 $^{*}$                                                     & 0.195 ± 0.036 $^{*}$                                                     \\
                          & mDCSRN $\dag$                                        & \textbf{47.72 ± 2.12 \textcolor{white}{$^{*}$}} & \textbf{0.962 ± 0.020 \textcolor{white}{$^{*}$}} & \textbf{0.012 ± 0.005 \textcolor{white}{$^{*}$}} & \textbf{0.184 ± 0.043 \textcolor{white}{$^{*}$}} \\ \hline
\multirow{2}{*}{UNet}     & 3D-cGAN \textcolor{white}{$\dag$} & 47.54 ± 1.98 $^{*}$                                                  & 0.961 ± 0.015 $^{*}$                                                    & 0.013 ± 0.003 $^{*}$                                                     & 0.192 ± 0.046 $^{*}$                                                     \\
                          & 3D-cGAN $\dag$                                       & \textbf{47.81 ± 2.03 \textcolor{white}{$^{*}$}} & \textbf{0.965 ± 0.016 \textcolor{white}{$^{*}$}} & \textbf{0.010 ± 0.003 \textcolor{white}{$^{*}$}} & \textbf{0.182 ± 0.048 \textcolor{white}{$^{*}$}} \\ \hline
\end{tabular}
}
\label{table_ablation_BDN}
\end{table}

We also apply SAN and RALS to various denoising networks by replacing the BDN in the framework with networks incorporating different architectures (such as ResNet \citep{he_resnet_2016}, DenseNet \citep{huang_densenet_2017}, and UNet \citep{ronneberger_unet_2015}). As shown in \autoref{table_ablation_BDN}, the integration of SAN and RALS leads to consistent improvements in all four evaluation metrics for three different denoising networks. Especially, the MAE within the lesion region is reduced by 0.024 for the ResNet-based BDN, 0.011 for the DenseNet-based mDCSRN, and 0.010 for the UNet-based 3D-cGAN. These results confirm that the two proposed components improve diagnostic image quality and demonstrate robust performance across different network architectures.

\begin{table}[!t] 
\caption{Ablation study on different domain knowledge. The best results are highlighted in bold. * Denotes results that are significantly different from the best results, based on a paired t-test with Bonferroni correction.}
\centering
\resizebox{0.5\textwidth}{!}{
\begin{tabular}{c|c|c|c|c}
\hline
Domain   Knowledge & PSNR↑                                                                & SSIM↑                                                                   & LPIPS↓                                                                     & MAE (Lesion)↓                                                           \\ \hline
$I^{L}_{d}$        & 48.13 ± 2.27 $^{*}$                                                  & 0.963 ± 0.017 $^{*}$                                                  & 0.008 ± 0.002 $^{*}$                                                    & 0.169 ± 0.054 $^{*}$                                                  \\
HFC                & 48.27 ± 2.32 $^{*}$                                                  & 0.964 ± 0.017 $^{*}$                                                  & 0.008 ± 0.003 $^{*}$                                                    & 0.168 ± 0.053 $^{*}$                                                  \\
$I^{SF}_{d}$       & \textbf{48.55 ± 2.28 \textcolor{white}{$^{*}$}} & \textbf{0.967 ± 0.014 \textcolor{white}{$^{*}$}} & \textbf{0.007 ± 0.003 \textcolor{white}{$^{*}$}} & \textbf{0.165 ± 0.052 \textcolor{white}{$^{*}$}} \\ \hline
\end{tabular}

}
\label{table_domain_knowledge}
\end{table}

\begin{table}[!t] 
\caption{Ablation study on different domain knowledge embedding methods. The best results are highlighted in bold. * Denotes results that are significantly different from the best results, based on a paired t-test with Bonferroni correction.}
\centering
\resizebox{0.5\textwidth}{!}{
\begin{tabular}{c|c|c|c|c}
\hline
Embedding Method       & PSNR↑                                                                & SSIM↑                                                                   & LPIPS↓                   & MAE (Lesion)↓                                                           \\ \hline
Single Embedding       & 48.22 ± 2.29 $^{*}$                                                  & 0.965 ± 0.017 $^{*}$                                                  & 0.008 ± 0.003          & 0.170 ± 0.050 $^{*}$                                                  \\
Hierarchical Embedding & \textbf{48.55 ± 2.28 \textcolor{white}{$^{*}$}} & \textbf{0.967 ± 0.014 \textcolor{white}{$^{*}$}} & \textbf{0.007 ± 0.003} & \textbf{0.165 ± 0.052 \textcolor{white}{$^{*}$}} \\ \hline
\end{tabular}
}
\label{table_hierarchy_embedding}
\end{table} 

\begin{table}[!t] 
\caption{Ablation study on $\mathcal{L}_{align}$. The best results are highlighted in bold. * Denotes results that are significantly different from the best results, based on a paired t-test with Bonferroni correction.}
\centering
\resizebox{0.5\textwidth}{!}{
\begin{tabular}{c|c|c|c|c}
\hline
Method                       & PSNR↑                                                                & SSIM↑                                                                   & LPIPS↓                                                                  & MAE (Lesion)↓                                                           \\ \hline
w/o    $\mathcal{L}_{align}$ & 48.38 ± 2.36 $^{*}$                                                  & 0.966 ± 0.014 $^{*}$                                                  & 0.008 ± 0.003 $^{*}$                                                  & 0.168 ± 0.053 $^{*}$                                                  \\
w/    $\mathcal{L}_{align}$  & \textbf{48.55 ± 2.28 \textcolor{white}{$^{*}$}} & \textbf{0.967 ± 0.014 \textcolor{white}{$^{*}$}} & \textbf{0.007 ± 0.003 \textcolor{white}{$^{*}$}} & \textbf{0.165 ± 0.052 \textcolor{white}{$^{*}$}} \\ \hline
\end{tabular}
}
\label{table_L_align}
\end{table}

 \textbf{(2) Ablation study on SAN.} We investigate the effectiveness of major components in SAN. \autoref{table_domain_knowledge} indicates that in comparison with the input $I^L_d$ and its high-frequency component (HFC) \citep{liu_promt_evp_2023}, the shallow feature $I_d^{SF}$ is a more effective form of domain knowledge to tune the model. Specifically, compared with $I^L_d$ and HFC, it achieves reductions in lesion MAE of 0.004 and 0.003, respectively, which correspond to improved lesion restoration. The comparison in \autoref{table_hierarchy_embedding} involves our hierarchical embedding method, utilizing distinct hierarchical embeddings for style modulation, against StyleGAN's single embedding approach \citep{karras_stylegan_2019,karras_stylegan2_2020}, which utilizes only the output embedding for style modulation, in all four evaluation metrics. In particular, it preserves lesions more effectively, achieving a lower lesion MAE of 0.005 than the single embedding approach. In \autoref{table_L_align}, our model with $\mathcal{L}_{align}$ outperforms the model trained without $\mathcal{L}_{align}$ by 0.17 dB in PSNR, 0.001 in SSIM, 0.001 in LPIPS, and 0.003 in lesion MAE, demonstrating the effectiveness of $\mathcal{L}_{align}$.

\begin{figure}[!t]
\centering
\includegraphics[width=0.48\textwidth]{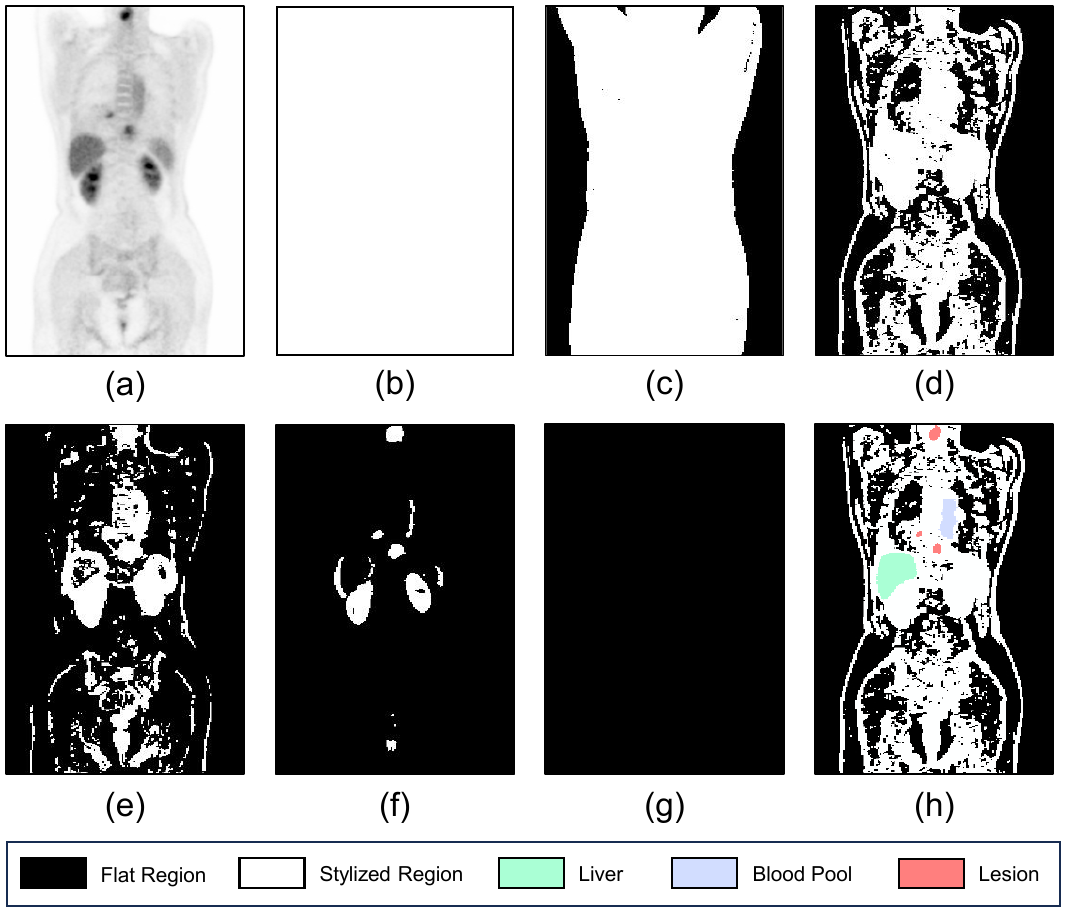}
\caption{Visualization of different masks. (a) Full-dose image. (b)-(g) Stylized region masks $M_{\delta}$ derived with varying thresholds: (b) $\delta=0$, (c) $\delta=0.0001$, (d) $\delta=0.001$, (e) $\delta=0.01$, (f) $\delta=0.1$, (g) $\delta=+\infty$. (h) Masks annotating three clinical ROIs (liver, blood pool, and lesion), overlaid on the stylized region mask with the final selected threshold $\delta=0.001$.} 
\label{fig_mask_and_roi}
\end{figure}

\textbf{(3) Ablation study on RALS.} We conduct an ablation study on the threshold $\delta$ used for identifying stylized regions. When $\delta=0$, RALS degenerates into GAN, as $\delta=0$ implies conducting adversarial learning across the entire image. When $\delta=+\infty$, RALS is disabled, as $\delta=+\infty$ indicates no region is selected for adversarial learning. \autoref{fig_mask_and_roi} shows an example of a stylized region with different threshold $\delta$. As observed, as $\delta$ increases, the size of the stylized region decreases. \autoref{table_ablation_rals} demonstrates that RALS utilizing these specified thresholds surpasses GAN in LPIPS, with $\delta=0.001$ yielding the best performance for RALS. This suggests that RALS, which employs adversarial learning in stylized regions, can more effectively learn and recover style information compared to the original GAN.

Furthermore, we evaluate their performance in specific clinical ROIs. It can be observed from \autoref{table_rals_clinical} that RALS reduces the MAE compared with GAN by 0.002 for the blood pool, 0.005 for the liver, and 0.004 for lesions. Additionally, by substituting the stylized region masks in RALS with manually annotated masks of specified clinical ROIs, we individually fine-tune the model on patches containing these three clinical ROIs, establishing an upper bound for RALS performance in each clinical region. The results demonstrate that utilizing stylized regions as guidance approaches the performance of methods using clinical annotations as guidance. This outcome is reasonable, as shown in \autoref{fig_mask_and_roi}. The final stylized region ($\delta=0.001$) strikes a good balance; it generally covers the three clinical ROIs and most other clinically important anatomical regions, while excluding easily recoverable flat regions.

\begin{table}[!t] 
\caption{Ablation study on RALS. The best results are highlighted in bold. * Denotes results that are significantly different from the best results, based on a paired t-test with Bonferroni correction.}
\centering
\resizebox{0.5\textwidth}{!}{
\begin{tabular}{c|c|c|c}
\hline
Method                                             & LPIPS↓                                 & Method                                       & LPIPS↓                                          \\ \hline
 w/o RALS ($\delta=+\infty$) &  0.011 ± 0.005 $^{*}$ &  RALS ($\delta=0.001$) &  \textbf{0.007 ± 0.003 \textcolor{white}{$^{*}$}} \\
 GAN ($\delta=0$)            &  0.009 ± 0.002 $^{*}$ &  RALS ($\delta=0.01$)  &  0.009 ± 0.003 $^{*}$          \\
 RALS ($\delta=0.0001$)      &  0.008 ± 0.002 $^{*}$ &  RALS ($\delta=0.1$)   & 0.010 ± 0.003 $^{*}$          \\ \hline
\end{tabular}

}
\label{table_ablation_rals}
\end{table}

\begin{table}[!t] 
\caption{Ablation study on RALS in clinical regions. Region denotes where the GAN training is conducted. The best results are highlighted in bold. * Denotes results that are significantly different from the best results, based on a paired t-test with Bonferroni correction.}
\centering
\resizebox{0.5\textwidth}{!}{
\begin{tabular}{cc|c|ccc}
\hline
\multicolumn{2}{c|}{\multirow{2}{*}{Method}}                                                                   & \multirow{2}{*}{Region} & \multicolumn{3}{c}{MAE↓}                                                                                                 \\ \cline{4-6} 
\multicolumn{2}{c|}{}                                                                                          &                         & \multicolumn{1}{c|}{Blood Pool}               & \multicolumn{1}{c|}{Liver}                    & Lesion                   \\ \hline
\multicolumn{2}{c|}{w/o RALS}                                                                                  & -           & \multicolumn{1}{c|}{0.105 ± 0.029 $^{*}$}          & \multicolumn{1}{c|}{0.110 ± 0.025 $^{*}$}          & 0.172 ± 0.047 $^{*}$          \\
\multicolumn{2}{c|}{GAN}                                                                                       & Entire Image           & \multicolumn{1}{c|}{0.100 ± 0.025 $^{*}$}          & \multicolumn{1}{c|}{0.104 ± 0.022 $^{*}$}          & 0.169 ± 0.048 $^{*}$          \\
\multicolumn{2}{c|}{RALS}                                                                                      & Stylized Region         & \multicolumn{1}{c|}{\textbf{0.098 ± 0.027 \textcolor{white}{$^{*}$}}}    & \multicolumn{1}{c|}{\textbf{0.099 ± 0.023 \textcolor{white}{$^{*}$}}}    & \textbf{0.165 ± 0.052 \textcolor{white}{$^{*}$}}    \\ \hline
\multicolumn{1}{c|}{\multirow{3}{*}{\begin{tabular}[c]{@{}c@{}}Upper\\ \\ Bond\end{tabular}}} & RALS & Blood Pool              & \multicolumn{1}{c|}{0.092 ± 0.025 \textcolor{white}{$^{*}$}} & \multicolumn{1}{c|}{—}                        & —                        \\
\multicolumn{1}{c|}{}                                                                         & RALS & Liver              & \multicolumn{1}{c|}{—}                        & \multicolumn{1}{c|}{0.093 ± 0.020 \textcolor{white}{$^{*}$}} & —                        \\
\multicolumn{1}{c|}{}                                                                         & RALS & Lesion              & \multicolumn{1}{c|}{—}                        & \multicolumn{1}{c|}{—}                        & 0.153 ± 0.053 \textcolor{white}{$^{*}$} \\ \hline
\end{tabular}
}
\label{table_rals_clinical}
\end{table}

\begin{table}[!t] 
\caption{Ablation study on the order of hyperparameters in tuning. The best results are highlighted in bold. * Denotes results that are significantly different from the best results, based on a paired t-test with Bonferroni correction.}
\centering
\resizebox{0.5\textwidth}{!}{
\begin{tabular}{cc|c|c}
\hline
\multicolumn{2}{c|}{Order}                                                                                                  & Configuration Changes & LPIPS↓                   \\ \hline
\multicolumn{1}{c|}{\multirow{4}{*}{Intra-group}} & data ($P$) → model ($N$ → $C$) → loss ($\beta$ → $\gamma$ →   $\delta$) & None                  & \textbf{0.007 ± 0.003  \textcolor{white}{$^{*}$}} \\
\multicolumn{1}{c|}{}                             & data ($P$) → model ($N$ → $C$)   → loss ($\gamma$ → $\delta$ → $\beta$) & None                  & \textbf{0.007 ± 0.003  \textcolor{white}{$^{*}$}} \\
\multicolumn{1}{c|}{}                             & data ($P$) → model ($C$ → $N$)   → loss ($\beta$ → $\gamma$ → $\delta$) & None                  & \textbf{0.007 ± 0.003  \textcolor{white}{$^{*}$}} \\
\multicolumn{1}{c|}{}                             & data ($P$) → model ($C$ → $N$)   → loss ($\gamma$ → $\delta$ → $\beta$) & None                  & \textbf{0.007 ± 0.003  \textcolor{white}{$^{*}$}} \\ \hline
\multicolumn{1}{c|}{\multirow{6}{*}{Inter-group}} & data → model → loss                                                     & None                  & \textbf{0.007 ± 0.003  \textcolor{white}{$^{*}$}} \\
\multicolumn{1}{c|}{}                             & data → loss → model                                                     & $\beta=0.0001$        & 0.008 ± 0.003 $^{*}$         \\
\multicolumn{1}{c|}{}                             & model → data → loss                                                     & None                  & \textbf{0.007 ± 0.003  \textcolor{white}{$^{*}$}} \\
\multicolumn{1}{c|}{}                             & model → loss → data                                                     & None                  & \textbf{0.007 ± 0.003  \textcolor{white}{$^{*}$}} \\
\multicolumn{1}{c|}{}                             & loss → data → model                                                     & $\beta=\delta=0.0001$ & 0.009 ± 0.003 $^{*}$         \\
\multicolumn{1}{c|}{}                             & loss → model → data                                                     & $\beta=\delta=0.0001$ & 0.009 ± 0.003 $^{*}$         \\ \hline
\end{tabular}

}
\label{table_hyper_params_order}
\end{table}

\begin{figure}[!t]
\centering
\includegraphics[width=0.48\textwidth]{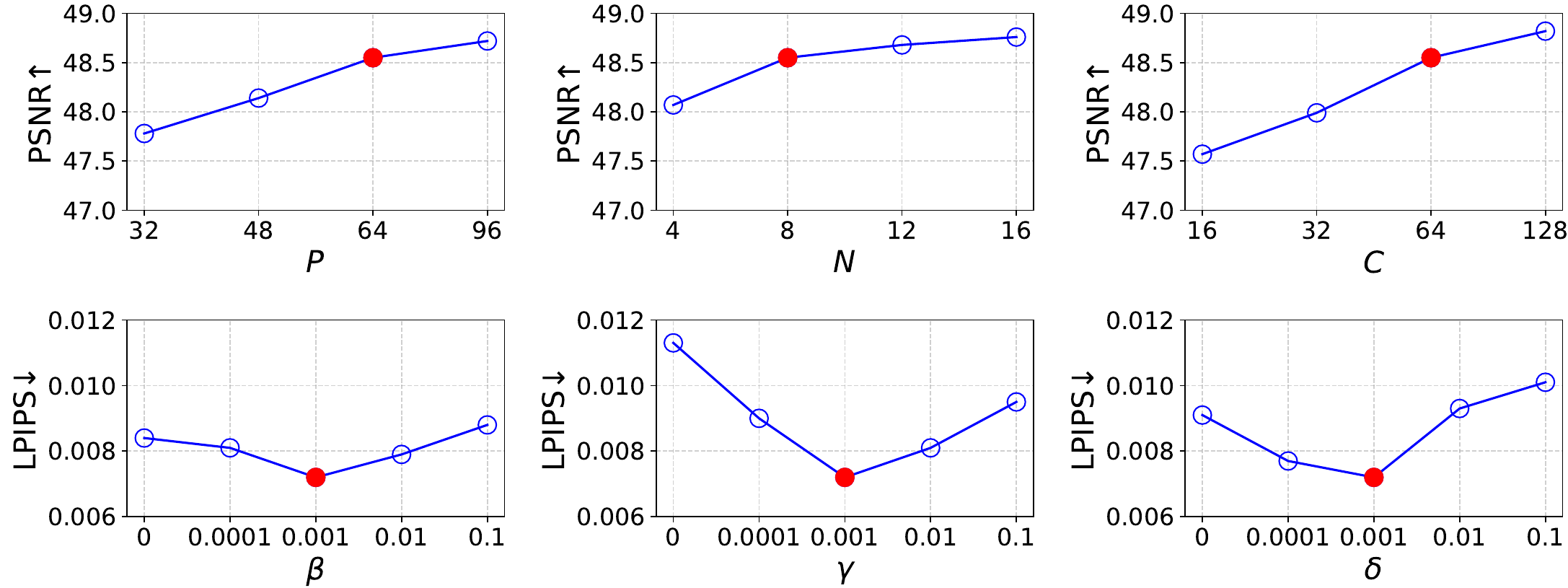}
\caption{Ablation study on hyperparameter. The red point marks the hyperparameter of the final configuration.} 
\label{fig_hyperparam}
\end{figure}

\textbf{(3) Ablation study on hyperparameter.} To demonstrate the impact of hyperparameters and validate the rationale behind our hyperparameters selection, we conduct ablation studies on several key hyperparameters: patch size $P$, number of residual blocks $N$, channel dimension $C$, loss balancing factors $\beta$ and $\gamma$, and the threshold $\delta$ that identifies the stylized region. The PSNR metric is used to evaluate $P$, $N$, and $C$ as these parameters have a greater influence on the image fidelity. The LPIPS metric is employed to assess $\beta$, $\gamma$, and $\delta$ since these parameters primarily affect image textures and details. As shown in \autoref{fig_hyperparam}, all six hyperparameters significantly impact model performance. The final selected values for each hyperparameter correspond to a turning point, beyond which model performance improves slowly or even degrades as the hyperparameter values increase. 

To evaluate the impact of hyperparameter tuning order, we conduct ablation studies by randomly shuffling sequences both within hyperparameter groups (intra-group) and across groups (inter-group). As shown in \autoref{table_hyper_params_order}, intra-group reordering does not affect the final configuration, suggesting that these hyperparameters within the same group are relatively independent. In contrast, inter-group shuffling introduces some variability. In particular, tuning loss-related hyperparameters too early consistently yields a suboptimal configuration. We attribute this to the large impact of data- and model-related hyperparameters, which form the overall optimization landscape. By comparison, loss-related hyperparameters primarily serve to fine-tune performance within that landscape and thus are more dependent on its prior configuration. These findings validate that our strategy of prioritizing data- and model-related hyperparameters before tuning loss-related parameters is effective.

\section{Discussion} 
In this paper, we propose the UniPET model for high-quality PET image denoising across different DRFs. The key innovation behind our method is the use of domain generalization techniques to learn domain-transferable features through style alignment. Specifically, we introduce a style alignment network (SAN) to align and recover different domain styles, ensuring both model generalization and high-quality denoising. In addition, to facilitate style recovery, we introduce a region-aware learning strategy (RALS) to guide the model's focus on stylized regions. We validate UniPET through extensive experiments on four datasets, which include both private clinical data and publicly available benchmarks. Our evaluation protocol rigorously tests performance across three challenging scenarios: (i) denoising under varying DRFs, (ii) out-of-distribution (OOD) data with unknown DRFs, and (iii) OOD data from an unknown center. Performance is quantified through three different evaluation perspectives: (i) quantitative metrics (PSNR and SSIM), (ii) perceptual quality measured by LPIPS and a reader score comparison, and (iii) clinical relevance assessed by SUV error on clinical ROIs, lesion detectability by radiologists, and lesion detectability by a classification model. The comprehensive experimental results consistently demonstrate UniPET’s superiority over existing methods, achieving state-of-the-art (SOTA) performance and generalizability.

However, our study still has some limitations. \textbf{From the perspective of model performance,} while our proposed UniPET achieves SOTA performance, small lesions in the synthesized images still exhibit a noticeable gap when compared to full-dose PET images. As shown in \autoref{fig_method_comparison_lesion}, \autoref{fig_method_comparison}, and \autoref{fig_method_comparison_Siemens_lesion}, the shape of the lesions in these synthesized images exhibits some degree of distortion relative to the full-dose images. This discrepancy is primarily due to the fact that small lesions make up a very small portion of the dataset, which makes it challenging for the model to accurately recover them and distinguish them from surrounding noise. In the future, we could follow the approach outlined in \citep{zhou_SGSGAN_2022} and incorporate auxiliary tasks, such as segmentation, to specifically enhance the recovery of small lesions. 

\begin{table}[t] 
\caption{Effect of the protective technique. The best results are highlighted in bold. * Denotes results that are significantly different from the best results, based on a paired t-test with Bonferroni correction.}
\centering
\resizebox{0.5\textwidth}{!}{
\begin{tabular}{c|ccc}
\hline
\multirow{2}{*}{Method} & \multicolumn{3}{c}{PSNR↑}                                                                                                            \\ \cline{2-4} 
                        & \multicolumn{1}{c|}{UPID-Base}               & \multicolumn{1}{c|}{UPID-OOD-DRF}                 & UPID-OOD-Center              \\ \hline
UniPET                  & \multicolumn{1}{c|}{\textbf{48.55 ± 2.28 \textcolor{white}{$^{*}$}}} & \multicolumn{1}{c|}{47.93 ± 2.28 $^{*}$}          & 43.61 ± 1.89 $^{*}$          \\
UniPET + FGSM           & \multicolumn{1}{c|}{48.36 ± 2.21 $^{*}$}          & \multicolumn{1}{c|}{\textbf{48.02 ± 2.22 \textcolor{white}{$^{*}$}}} & \textbf{43.83 ± 1.95 \textcolor{white}{$^{*}$}} \\ \hline
\end{tabular}
}
\label{table_discussion_FGSM}
\end{table}

\textbf{From the perspective of model generalizability,} although UniPET incorporates domain generalization techniques, it still faces risks when tested on out-of-distribution (OOD) data. As shown in \autoref{fig_ood_center}, the synthesized result of UniPET on the UPID-OOD-Center dataset exhibits some degree of over-smoothing, and the contrast of one lesion is not adequately recovered. This suggests that the model has inherent vulnerabilities that cannot be addressed simply by learning from the limited training data. To mitigate this issue, protective techniques like the fast gradient sign method (FGSM) \citep{goodfellow2014FGSM} could be introduced. FGSM helps identify the model's vulnerabilities during training by generating adversarial samples through intentional perturbations to the input, thereby maximizing the model's loss. The model is then trained to minimize this loss, which improves its ability to handle adversarial attacks and address these vulnerabilities. This approach not only enhances the model's robustness to adversarial attacks but also improves its generalizability to potentially unseen data and unknown domains. We apply FGSM to our proposed UniPET, and the experimental results are presented in \autoref{table_discussion_FGSM}. As observed, while the model’s performance slightly decreases on the UPID-Base dataset after applying FGSM, its performance improves on the OOD datasets, UPID-OOD-DRF and UPID-OOD-Center. This indicates that incorporating protective techniques like FGSM can enhance model generalizability. In the future, we may explore more advanced protective techniques to further improve model robustness and generalizability.

\textbf{From the perspective of model evaluation,} while we employ a range of evaluation measures—quantitative, perceptual, and clinical—to show that our proposed UniPET outperforms existing SOTA methods, a key concern remains regarding the clinical acceptability of these metrics. Specifically, it is still unclear to what extent these evaluation measures align with real-world clinical standards and whether the model itself can be deemed clinically acceptable. To address this, future work could focus on evaluating the model's performance through a large-scale benchmark based on real clinical patient data. This would provide more direct insights into the model’s practical utility, ensuring its performance aligns with clinical expectations and validating its readiness for clinical deployment.

\section{Conclusion} 
In this paper, we focus on the task of universal PET image denoising, aiming to recover low-dose PET images across different DRFs. To realize this, we harness the concept of domain generalization and propose a universal PET image denoising network (UniPET). Specifically, UniPET consists of a pre-trained base denoising network (BDN) for coarse-grained denoising, a style alignment network (SAN) for aligning and recovering fine-grained styles across different DRFs, and a region-aware learning strategy (RALS) to direct the model's focus toward learning stylized regions. Comprehensive experiments demonstrate that our proposed UniPET exhibits comparable performance to DRF-specific models and achieves state-of-the-art performance in universal PET image denoising quantitatively, perceptually, and clinically.

\section*{Declaration of Competing Interest}

The authors declare that they have no known competing financial interests or personal relationships that could have appeared to influence the work reported in this paper.

\section*{Acknowledgments}

This work is supported by the National Natural Science Foundation in China under Grant 62371016, U23B2063, 62022010, and 62176267, the Bejing Natural Science Foundation Haidian District Joint Fund in China under Grant L222032, the Beijing hope run special fund of cancer foundation of China under Grant LC2018L02, the Fundamental Research Funds for the Central University of China from the State Key Laboratory of Software Development Environment in Beihang University in China, the 111 Proiect in China under Grant B13003, the SinoUnion Healthcare Inc. under the eHealth program, the high performance computing resources at Beihang University. ChatGPT is used only for checking grammar. All scientific content and references are prepared and verified by the authors.

\bibliographystyle{model2-names.bst}\biboptions{authoryear}
\bibliography{refs}

\end{document}